\documentclass{article}

\usepackage[eandd, final]{neurips_2026}

\usepackage[utf8]{inputenc}
\usepackage[T1]{fontenc}
\usepackage{hyperref}
\hypersetup{hidelinks}
\usepackage{url}
\usepackage{booktabs}
\usepackage{amsfonts}
\usepackage{amsmath}
\usepackage{nicefrac}
\usepackage{microtype}
\usepackage{xcolor}
\usepackage{graphicx}
\usepackage{array}
\usepackage{CJKutf8}
\usepackage[most]{tcolorbox}

\title{AVID: A 5T fMRI Dataset for Benchmarking Auditory-induced Visual Mental Imagery Decoding}

\author{Shiqi Shen$^{1,2,3,*}$ \quad
  Shurui Li$^{1,*}$ \quad
  Xilin Zhang$^{2,\dagger}$ \quad
  Yuanning Li$^{1,3,\dagger}$
  \\
  \\
  $^{*}$Equal contribution \qquad
  $^{\dagger}$Corresponding authors
  \\
  $^{1}$ShanghaiTech University \qquad
  $^{2}$South China Normal University \qquad
  $^{3}$XS Vision
  \\
  \texttt{2023010281@m.scnu.edu.cn},
  \texttt{lishr2022@shanghaitech.edu.cn}
  \\
  \texttt{xlzhang@m.scnu.edu.cn},
  \texttt{liyn2@shanghaitech.edu.cn}
}

\begin{document}

\maketitle

\begin{abstract}
Visual mental imagery (VMI) decoding uses brain activity to reconstruct internally generated visual scenes, providing a potential pathway for externalizing internal experiences and developing neuroprosthetic communication interfaces. However, progress in this field is limited by the lack of dedicated, training-scale VMI datasets. 
Current VMI paradigms often rely on visual prompts or short-term memory images, introducing visual and mnemonic confounds that make it difficult to disentangle neural activity associated with endogenous image construction from signals driven by prior or concurrent visual input.
Furthermore, existing VMI datasets offer limited descriptive control over individual visual attributes, making it difficult to systematically assess fine grained visual details.
To address these gaps, we introduce AVID, the first dedicated training-scale 5.0 Tesla functional magnetic resonance imaging (fMRI) resource for auditory-induced VMI decoding. 
Acquired with high-field 5T fMRI, AVID comprises approximately 67 hours of dense recordings from nine participants. The dataset is uniquely structured into two complementary splits: a scene-level complex split using naturalistic Mandarin descriptions and a controlled simple split with explicitly specified visual attributes. We further formalize a standardized evaluation protocol that decouples source-image consistency from semantic alignment, allowing for a more nuanced assessment of imagery reconstruction. 
Baseline qualitative and quantitative results show that current reconstruction pipelines remain limited on AVID, although the most successful reconstructions preserve recognizable scene level semantics from spoken descriptions. By combining dedicated training-scale neural recordings and standardized evaluation resources, AVID provides both a data foundation for imagery-specific model development and a benchmark for evaluating VMI decoding with substantially reduced sensory contamination.

\end{abstract}

\section{Introduction}

Visual mental imagery (VMI) refers to the ability to internally construct visual scenes in the absence of sensory input. This capability supports core cognitive functions such as planning, memory, and language grounding. Decoding the neural representations underlying VMI offers a potential pathway toward externalizing internal visual experiences, with implications for neuroprosthetic systems and communication interfaces.

Despite recent advances in neural decoding and brain-to-image reconstruction~\citep{takagi2023high,ozcelik2023natural,scotti2023reconstructingmindseyefmritoimage,lu2023minddiffuser,scotti2024mindeye2,wang2024mindbridge}, most progress has been driven by stimulus-based visual perception. In contrast, reconstructing imagined content remains substantially more challenging~\citep{kneeland2025nsd}. Mental imagery is not time-locked to an external stimulus and typically produces weaker and more variable neural responses, with substantial inter-subject variability in imagery strategies and vividness. Moreover, VMI is an endogenous process that can be elicited through non-visual modalities such as language or auditory input, further complicating both modeling and evaluation.

Existing VMI studies remain limited in both scale and experimental design. Many paradigms rely on memorized images or visually presented prompts~\citep{miyawaki2008visual,dijkstra2017vividness,naselaris2015voxel,shen2019deep}, introducing confounds such as residual perceptual activity and short-term or working-memory replay. Although some work has explored auditory cues~\citep{ganis2004brain,senden2019reconstructing}, these approaches are typically restricted to isolated words or low-dimensional categories and do not capture the compositional structure of natural scenes. As a result, current resources provide limited traction for studying scene-level mental imagery under reduced sensory contamination.

More fundamentally, evaluating VMI decoding raises challenges that are not present in perception-based reconstruction. Unlike perception, imagined content does not have a unique pixel-level ground truth, making direct reconstruction evaluation ill-posed. Standard evaluation protocols, which assume a well-defined visual target, therefore introduce a mismatch between what is measured and what the task requires. In addition, existing benchmark designs often conflate perception, memory, and imagery, making it difficult to isolate endogenous image construction from residual sensory or mnemonic processes.

We therefore treat auditory-induced VMI decoding explicitly as an evaluation problem rather than only a reconstruction problem. To this end, we introduce a visual-free, auditory-induced paradigm and collect the Auditory-induced Visual Imagery Dataset (AVID), a 5.0 Tesla fMRI dataset acquired under two complementary stimulus regimes. The complex split uses natural-scene descriptions to probe scene-level mental imagery, while the simple split uses controlled descriptions with explicitly specified attributes such as identity, color, count, alphanumeric content, and position. Together, these splits make it possible to distinguish variation due to open-ended scene ambiguity from incomplete recovery of verbally specified information.

We further define a standardized evaluation protocol that jointly considers source image consistency and semantic alignment with the spoken description. The protocol evaluates whether reconstructions remain anchored to the source image used to derive the spoken description, while also capturing the verbally specified content conveyed to the participant. We report baseline results using four representative visual decoding pipelines, LDM~\citep{takagi2023high}, MindEye1~\citep{scotti2023reconstructingmindseyefmritoimage}, MindEye2~\citep{scotti2024mindeye2}, and MindBridge~\citep{wang2024mindbridge}, treating them as probes of the benchmark rather than as claims of model novelty. 

Our contributions are as follows.
\begin{itemize}
  \item We introduce a visual-free, auditory-induced VMI paradigm designed to reduce visual contamination and short-term or working-memory replay confounds.
  \item We release AVID, the first training-scale 5T VMI fMRI dataset with a scene-level complex split and a controlled simple split with explicitly specified attributes.
  \item We define a standardized evaluation protocol for auditory-induced VMI decoding and provide baseline evaluation results under this protocol.
\end{itemize}

\section{Related work}

\subsection{fMRI image reconstruction from visual perception}

The development of deep learning in computer vision \citep{krizhevsky2012imagenet,dosovitskiy2020image,liang2021swinir,wu2025lightqanet} has changed the scope of neural decoding. Earlier fMRI decoding methods relied mainly on linear mappings, sparse priors, or Bayesian estimation \citep{kay2008identifying,naselaris2009bayesian,miyawaki2008visual,kamitani2005decoding}. These methods established important links between brain activity and low-level visual features or discrete semantic categories, but they were not intended to reconstruct the texture, layout, and spatial structure of natural images.

Recent fMRI based image reconstruction has been driven largely by visual perception datasets such as NSD \citep{allen2022massive} and by decoders that map brain activity into the latent spaces of modern generative models \citep{takagi2023high,scotti2023reconstructingmindseyefmritoimage,scotti2024mindeye2}. These methods provide strong baselines and useful starting points for our benchmark, but they also highlight a key distinction. In visual perception, the reconstruction target is an externally presented image. In auditory induced mental imagery, the target is internally generated and has no unique pixel level reference. This shift changes not only the decoding problem but also the assumptions under which reconstruction quality is measured.

Cognitive neuroscience also cautions against treating mental imagery as a direct replay of perception~\citep{wadia2026shared,pearson2015mental}. VMI recruits a distributed network that partly overlaps with perception, while differing in spatial precision, response variability, and the relative involvement of early visual and higher order regions~\citep{pearson2019human,spagna2021visual,li2024enhancing}. These differences are important for benchmark design. If imagery responses are weaker, less spatially stable, and more strongly mediated by semantic content than perception responses, then image similarity alone may fail to capture meaningful recovery of imagined content.

\subsection{Mental imagery benchmarks}

Existing VMI datasets are often constrained by the way imagery is elicited. Memory-based paradigms require participants to view or memorize specific images before scanning \citep{miyawaki2008visual,dijkstra2017vividness,naselaris2015voxel,kneeland2025nsd}. Such designs are valuable for studying the overlap between perception, memory, and imagery, but they make it difficult to separate endogenous image construction from residual visual traces or short-term/working-memory replay. Text-cued paradigms reduce reliance on memorized images, but on-screen instructions can still introduce visual input during the task \citep{hayashi2014visual,anderson2015reading,albers2013shared}. These constraints motivate benchmark designs in which imagery is specified without concurrent visual presentation.

The closest existing benchmark is NSD-Imagery, which studies mental imagery in NSD participants using a relatively small imagery sample built on the much larger NSD resource \citep{kneeland2025nsd}. This setting is well suited for testing how decoders trained on visual perception data transfer to imagery, including cases cued by visually presented words or text. Its results also suggest that such transfer remains limited.

\section{Acquisition Paradigm and Dataset Construction}
\label{paradigm and dataset}

\subsection{Overview}

AVID~\citep{avid_dataset2026} was designed to study visual mental imagery elicited without concurrent visual input. Its acquisition protocol therefore focuses on two requirements. First, the imagery cue should specify sufficiently rich visual content while avoiding visual presentation during the imagery period. Second, the resulting fMRI recordings should support controlled analyses of both natural scene imagery and explicitly specified visual attributes. To meet these requirements, we used spoken descriptions, eyes closed imagery, dense within participant sampling, and two complementary stimulus splits.

\begin{figure}[!t]
    \centering
    \includegraphics[width=1\linewidth]{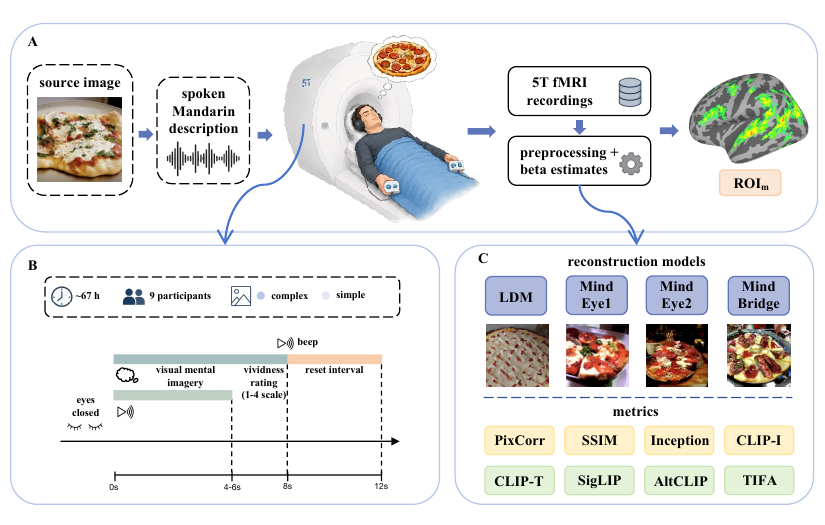}
    \caption{Overview of our benchmark.
    (A) Source images are converted into spoken Mandarin descriptions, which participants hear during eyes-closed imagery in the 5T scanner. The fMRI data are preprocessed to estimate trial-wise responses and define ROI-based neural inputs for decoding. (B) AVID includes approximately 67 hours of recordings from 9 participants across complex scene-level and controlled attribute splits, using a trial structure with an auditory cue, imagery period, vividness rating, and reset interval. (C) Decoders are evaluated on fixed AVID splits, with reconstructions scored by source-image consistency and spoken-description alignment.}
    \label{fig:benchmark}
\end{figure}

\subsection{Participants}
\label{participants}

Nine healthy adults aged 18 to 25 were recruited from the student population. All participants were healthy native Mandarin speakers with normal or corrected-to-normal vision and normal hearing. The study was approved by the Institutional Review Board of ShanghaiTech University (IRB protocol number: Q2023-007), and all participants provided written informed consent in accordance with institutional guidelines. Participants were also informed of the long-term nature of the study prior to enrollment and were naive to the specific experimental hypotheses. Participants were financially compensated at a rate of 100 RMB per hour. They were informed of their right to withdraw from the study at any time without penalty and were also made aware that their data would be used for scientific research and, following anonymization, would be shared publicly.

Participants were pre-screened using the Vividness of Visual Imagery Questionnaire (VVIQ) \citep{marks1973visual}, with scores ranging from 52 to 81, indicating relatively vivid self-reported imagery. All participants completed the main mental imagery task and a functional localization task described in Section~\ref{Localizer}. The simple split was collected from seven of these participants under the same recruitment criteria and task instructions. More details are provided in Appendix~\ref{app:ExperimentProcedure}.

\subsection{Visual-free auditory-induced paradigm}
\label{Paradigm}

In AVID, participants performed an event-related mental imagery task (Figure~\ref{fig:benchmark}), in which each trial began with a spoken description. The auditory cues are spoken Mandarin descriptions that specify visual content without presenting visual stimuli. The description specified the content to be imagined without presenting a visual prompt during the imagery period. Participants kept their eyes closed and formed a static visual scene from an observer perspective, rather than imagining themselves acting within the scene. This instruction was intended to reduce variability associated with self representation, motor planning, and egocentric spatial updating.

Each trial lasted 12 s. The first 8 s served as the imagery window. The spoken description was presented at trial onset, and participants maintained the elicited image throughout this period. A 250 ms auditory beep marked the transition to a 4 s reset interval before the next trial. Participants rated imagery vividness on a 4 point scale using an MR compatible button box. Additional implementation details on run and session organization are provided in Appendix~\ref{app:ExperimentProcedure}.

\subsection{Stimulus splits}

\paragraph{Complex split.}

The complex split uses scene-level spoken descriptions derived from processed image--caption stimulus pairs from the Caption Scene Dataset \citep{li2026large}, which were originally based on COCO-CN, a Mandarin image caption collection \citep{li2019coco}. Captions were synthesized into 4 to 6 s audio clips using Google Text-to-Speech, with each caption containing 12 to 20 Mandarin characters. We curated 800 unique complex stimuli, and each stimulus was presented twice to each participant in pseudo-randomized order. More details about stimuli preparation are provided in Appendix~\ref{app:Stimuli_preparation}

\paragraph{Simple split.}
The simple split contains 400 in-house image-caption stimulus pairs designed to make a small set of visual attributes explicit, including object identity, alphanumeric content, color, count, and relative position. The spoken descriptions describe sparse configurations on a neutral background, reducing semantic ambiguity while preserving the same spoken description imagery procedure used in the complex split. Each stimulus was repeated twice for each participant. 

The two splits are complementary rather than hierarchical. The complex split probes whether decoders can recover scene level semantic structure from natural descriptions, where multiple visual realizations are plausible. The simple split narrows this ambiguity by explicitly specifying a small number of inspectable attributes. It therefore provides a controlled setting for diagnosing whether reconstruction errors reflect failures to recover information that was unambiguously given to the participant. Table~\ref{tab:dataset_stats} summarizes the stimulus and trial counts for the two splits.

\begin{table}[t]
  \caption{Summary of AVID dataset statistics.}
  \label{tab:dataset_stats}
  \centering
  \begin{tabular}{lccccc}
    \toprule
    Split & Participants & Stimuli & Total trials & Task time (h) & Repetitions \\
    \midrule
    Complex & 9 & 800 & 14,400 & 48.0 & 2 \\
    Simple & 7 & 400 & 5,600 & 18.7 & 2 \\
    Total & 9 & 1,200 & 20,000 & 66.7 & 2 \\
    \bottomrule
  \end{tabular}
\end{table}

\subsection{Functional localizer and ROI definitions}
\label{Localizer}

The auditory-induced imagery paradigm engages speech perception, linguistic comprehension, and mental image construction within the same task. AVID therefore includes a functional localizer to dissociate these components and define corresponding ROI sets for benchmark inputs. The localizer follows the logic of individual-subject language localizers \citep{fedorenko2024language} and includes three auditory conditions: Mandarin sentences, non-word sequences, and non-linguistic acoustic controls.

AVID provides ROI definitions at three levels. A noise-ceiling analysis \citep{allen2022massive} over repeated mental imagery trials estimates response reliability and identifies task-responsive vertices, denoted as $ROI_m$. The contrast between non-word sequences and non-linguistic acoustic controls identifies speech-responsive vertices, denoted as $ROI_s$. The contrast between sentences and non-word sequences identifies language-selective vertices, denoted as $ROI_l$. These masks make the neural input definition part of the benchmark specification, allowing all baselines to be evaluated on the same pre-defined vertex sets rather than selecting inputs in a model-dependent way.

\subsection{MRI data acquisition and preprocessing}

Functional MRI data were acquired on a 5T MRI scanner. Functional images were collected with TR\,=\,1.5\,s, 2.0\,mm isotropic voxels, and an anterior to posterior phase-encoding direction, with matched posterior to anterior images acquired for distortion correction. High-resolution T1-weighted anatomical images were obtained before fMRI acquisition at 0.7\,mm isotropic resolution. Standard preprocessing included slice-timing correction, distortion correction, motion correction, anatomical alignment and one-step resampling. Additional details are provided in the Appendix~\ref{app:MRIacquisition} and~\ref{app:fMRIpreprocessing}.

\subsection{Quality control and neural signal validation}
We performed quality control at multiple stages to establish the quality and reliability of AVID. Temporal signal-to-noise ratio (tSNR) was computed from raw fMRI time series to assess acquisition quality (Appendix~\ref{app:tSNR}). Repeat-based noise-ceiling analyses assessed the reliability of trial-wise neural responses (Appendix~\ref{app:nc}). We also examined how neural reliability varied with self-reported imagery vividness (Appendix~\ref{app:vividness_relationship}). Neural signal validation further assessed decodable task-related information using complementary decoding analyses with shuffled-fMRI and matched-Gaussian controls (Appendix~\ref{app:neural_validation}). Together, these assessments support the signal quality, repeatability, and neural informativeness of AVID, providing a reliable foundation for auditory-induced visual imagery research.

\section{Evaluation protocol}
\label{sec_evaluation_protocol}

\subsection{Problem formulation}

Given fMRI activity recorded during an auditory induced VMI trial, a reconstruction pipeline generates multiple candidate images intended to reflect the participant's imagined content. Unlike visual perception reconstruction, the imagined content does not have a unique pixel level target. The source image associated with the spoken description provides a reference for constructing the spoken description, supervising the reconstruction model, and computing source-image-based metrics. However, it should not be interpreted as the participant's exact mental image.

Therefore, the protocol  evaluates reconstructions along two complementary axes. Source image consistency measures whether a reconstruction remains visually anchored to the source image from which the spoken description was derived. Spoken-description semantic alignment measures whether a reconstruction preserves the content that was verbally specified to the participant. Reporting these axes separately makes it possible to distinguish visual resemblance to the original source image from semantic recovery of the spoken description.

\subsection{Evaluation splits and sampling policy}

For the baseline evaluations, we use pre-specified stimulus-level splits within each stimulus split. The released split files define which stimuli are used for training and which are held out for testing, ensuring that fMRI responses to the same stimulus do not appear in both sets. This setting evaluates reconstruction from responses to held-out stimuli within the AVID stimulus distribution. However, the split is not designed to enforce strict semantic disjointness between training and test sets; semantically related descriptions or source images may still appear across the two sets.

In addition to the primary benchmark setting, we provide three baseline training settings to characterize common use cases of AVID: single-participant training, leave-one-participant-out transfer training, and multi-participant joint training. These settings are intended as reference baselines rather than mandatory components of the evaluation protocol. They cover subject-specific decoding from an individual’s own data, transfer to a held-out participant using data from other participants, and shared model training across participants. Across these baselines, we use the same released data partitions, neural input definitions, and metric suite. Each reconstruction pipeline retains its own generation and candidate-selection procedure, and the resulting outputs are evaluated with the same source-image consistency and spoken-description alignment metrics. 

\subsection{Neural input settings}

To keep model comparisons tied to the same neural evidence, all reconstruction pipelines operate on shared AVID fMRI inputs prepared by the benchmark pipeline. The inputs consist of trial wise beta responses after the preprocessing, repeat reliability analysis, and ROI definition steps described in Section~\ref{Localizer}. The primary neural input uses vertices in \(ROI_m \setminus ROI_s\), removing speech responsive vertices while retaining task responsive activity that may support VMI reconstruction.

Unlike the inputs commonly used for visual perception decoding, which provide broad coverage of visual cortex, AVID's primary input is distributed across frontoparietal, sensorimotor, temporal, and language-selective regions. Early visual cortex (EVC) is sparsely represented: EVC account for approximately 4.5\% and 5.8\% of the selected vertices in the complex and simple splits, respectively.

\subsection{Metrics}
The metric suite is designed to remain comparable with perception-based fMRI reconstruction work while reflecting the language-specified nature of auditory-induced VMI. We use spoken-description semantic alignment as the primary criterion for semantic recovery, and source-image consistency as a secondary diagnostic of source-compatible visual structure. The two groups are reported separately rather than combined into a single aggregate score.

\paragraph{Source-image consistency.}
Source-image consistency metrics evaluate whether reconstructions remain visually anchored to the source image associated with the spoken description. We report PixCorr and SSIM as direct image-comparison metrics, and Inception and CLIP-I as feature-based two-way identification accuracies. These metrics enable comparison with perception-based reconstruction work, but are not interpreted as direct measurements of the participant's exact imagery.

\paragraph{Spoken-description semantic alignment.}
Spoken-description semantic alignment metrics evaluate whether reconstructions preserve the content conveyed by the spoken description. We report CLIP-T~\citep{radford2021learning}, SigLIP~\citep{zhai2023sigmoid}, and AltCLIP~\citep{chen2023altclip} as image-text two-way identification accuracies, and TIFA~\citep{hu2023tifa} as a question-based evaluation of entities, attributes, and relations. Detailed metric definitions are provided in Appendix~\ref{app:metrics}.

\paragraph{Attribute-level recovery.}
For the simple split, the controlled experimental design allows us to additionally evaluate reconstruction accuracy for three attributes: color, spatial position, and visual identity. We use a frozen Qwen3-VL-8B model as an evaluator to answer closed-set questions about the reconstructed images. We report class-balanced accuracy for each attribute to account for uneven reference-label distributions. Evaluation details and evaluator validation are provided in Appendix~\ref{app:simple_attribute}.

\section{Baseline Methods}
\label{baseline}

We evaluate four representative fMRI-to-image reconstruction methods, LDM, MindEye1, MindEye2, and MindBridge, as baseline references for AVID. These methods cover three common training settings: LDM and MindEye1 are evaluated as single-participant models, MindEye2 is evaluated in a leave-one-participant-out transfer setting, and MindBridge is evaluated as a multi-participant joint model. Together, these baselines characterize how AVID supports subject-specific decoding, cross-participant transfer, and shared model training.

Our goal is not to introduce a new decoding architecture, but to examine how established reconstruction methods behave under auditory-induced VMI. To make the comparison faithful to prior work, we preserve each method’s original training and reconstruction setting whenever possible, including its model-selection, generation, and candidate-selection procedures. All methods are evaluated on the released AVID splits, predefined neural input definitions, and shared metric suite.

For AVID, the required adaptations are limited to data loading, neural input formatting, split handling, and evaluation with the released metrics. We do not introduce additional architectural changes or tune methods separately for individual metrics. Implementation details that are needed for reproducibility are provided in Appendix~\ref{app:baseline_implementation}.

\section{Results}
\label{Results}

We evaluate LDM, MindEye1, MindEye2, and MindBridge on the complex and simple splits using the primary neural input $ROI_m \setminus ROI_s$. This setting removes speech responsive vertices while retaining task responsive activity used for reconstruction. We report qualitative reconstructions together with two groups of metrics: source-image consistency and spoken-description alignment. As a stricter auditory-language control, we also evaluate $ROI_m \setminus (ROI_s \cup ROI_l)$; retained performance suggests decoding beyond speech- and language-selective regions (Appendix~\ref{app:sensitivity analysis}).

\subsection{Qualitative reconstructions}

\begin{figure}[t]
    \centering
    \includegraphics[width=1\linewidth]{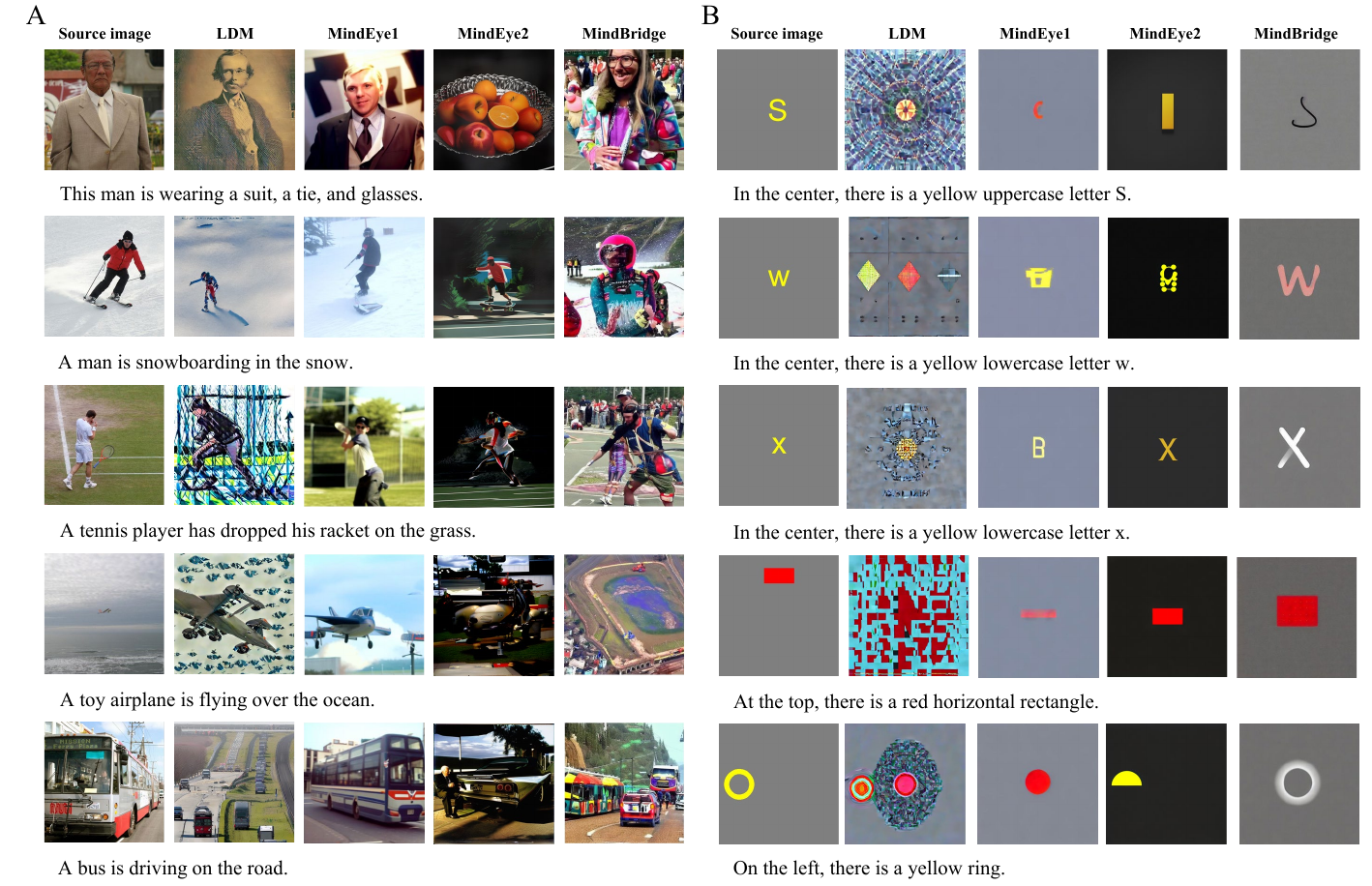}
    \caption{Qualitative comparison of mental imagery generation results. (A) Examples from the complex stimuli split. (B) Examples from the simple stimuli split. For each stimulus, we compare the source image with reconstructions generated by LDM, MindEye1, MindEye2, and MindBridge. The results illustrate how different methods recover semantic content and visual structure across natural images and simple shape stimuli.}
    \label{fig:ReconstructionFigure}
\end{figure}

Figure~\ref{fig:ReconstructionFigure} shows representative reconstructions from the complex and simple splits. MindEye1 gives comparatively stable reconstructions in terms of semantic coverage and overall scene completeness. Its outputs often preserve the dominant object category and coarse scene context, even when the precise layout or local details differ from the source image. LDM sometimes produces visually rich images with strong texture and local detail, but this texture can dominate the reconstruction and does not always correspond to the described content. MindBridge tends to introduce additional semantic elements that are not clearly supported by the description, and its reconstructions are visually less stable in several examples. MindEye2 is less consistent in semantic alignment on this split, often producing images that preserve only weak or partial correspondence to the spoken description. When a description contains several semantic elements, some methods capture only a subset of the described content or shift toward related but mismatched content. This behavior is consistent with the design of the complex split, where the source image anchors stimulus construction and source-image-based evaluation without serving as the participant’s exact imagined image.

The simple split complements this view. MindBridge appears comparatively stronger in the shown examples, often preserving the semantic identity of the target object and its approximate position. LDM is less reliable in semantic alignment for these sparse stimuli, but it sometimes predicts content at multiple plausible locations and can recover the correct color. Although visually sparse, it also contains multiple semantic attributes specified in the descriptions.

The reconstructions are not uniformly better than those in the complex split. Since these attributes are explicitly specified, such deviations are less plausibly attributed to open-ended variation between an imagined scene and its source image, and more directly suggest incomplete attribute recovery. Across both splits, the examples show that current reconstruction methods can recover recognizable content in favorable cases, but do not yet recover all specified semantics and attributes reliably.

\subsection{Complex split}

Table~\ref{tab_complex_results} reports the quantitative results on the complex split, which is the primary setting for evaluating scene-level visual imagery elicited from spoken natural descriptions. 
MindEye1 achieves the highest mean scores on CLIP-I, CLIP-T, SigLIP, and AltCLIP, whereas MindBridge obtains the highest mean Inception score. LDM also remains above chance on several semantic metrics. These results indicate that single-participant models can extract stable visual and semantic information from auditory-induced imagery responses.

MindBridge achieves mean performance close to MindEye1 on several feature- and text-based identification metrics, indicating that joint training across participants can support reconstruction within the training cohort. However, MindEye2, evaluated using leave-one-participant-out pretraining followed by target-participant fine-tuning, shows lower performance on these metrics. The remaining gap is therefore concentrated in the current transfer baseline, highlighting cross-participant transfer and adaptation as important directions for improving auditory-induced imagery decoding on AVID.

The low PixCorr values across all methods further show that pixel-level recovery of the source image is not an appropriate expectation for auditory-induced VMI. This is consistent with the design of the complex split, where the source image serves as a visual reference for constructing the spoken description and anchoring part of the evaluation, but not as the participant’s unique imagined image. A reconstruction may therefore differ from the exact source-image composition while still preserving the core entities or relations conveyed by the spoken description. These results support reporting source-image consistency and spoken-description alignment separately, rather than treating VMI reconstruction as a standard pixel-level image reconstruction problem.

\begin{table}[t]
  \caption{Complex split baseline results. Metrics are grouped by source image consistency and spoken description alignment.}
  \label{tab_complex_results}
  \centering
  \begin{tabular*}{\linewidth}{@{\extracolsep{\fill}}lcccccccc@{}}
    \toprule
    & \multicolumn{4}{c}{Image-image} & \multicolumn{4}{c}{Image-text} \\
    \cmidrule(lr){2-5} \cmidrule(lr){6-9}
    Method & PixCorr & SSIM & Inception & CLIP-I & CLIP-T & SigLIP & AltCLIP & TIFA \\
    \midrule
    LDM & 0.038 & 0.185 & 59.09\% & 59.61\% & 66.81\% & 68.54\% & 67.34\% & \textbf{32.56\%} \\
    MindEye1 & \textbf{0.056} & \textbf{0.291} & 65.14\% & \textbf{65.94\%} & \textbf{67.83\%} & \textbf{70.63\%} & \textbf{69.83\%} & 30.27\% \\
    MindEye2 & 0.034 & 0.262 & 55.75\% & 57.81\% & 57.95\% & 59.08\% & 58.82\% & 24.34\% \\
    MindBridge & 0.040 & 0.170 & \textbf{65.22\%} & 65.80\% & 67.66\% & 69.20\% & 69.61\% & 32.25\% \\
    \bottomrule
  \end{tabular*}
\end{table}

\begin{table}[t]
  \caption{Simple split baseline results. Metrics are grouped by source image consistency and spoken description alignment.}
  \label{tab_simple_results}
  \centering
  \begin{tabular*}{\linewidth}{@{\extracolsep{\fill}}lcccccccc@{}}
    \toprule
    & \multicolumn{4}{c}{Image-image} & \multicolumn{4}{c}{Image-text} \\
    \cmidrule(lr){2-5} \cmidrule(lr){6-9}
    Method & PixCorr & SSIM & Inception & CLIP-I & CLIP-T & SigLIP & AltCLIP & TIFA \\
    \midrule
    LDM & -0.001 & 0.636 & 54.07\% & 51.75\% & 55.59\% & \textbf{57.16\%} & \textbf{57.25\%} & \textbf{48.92\%} \\
    MindEye1 & -0.028 & \textbf{0.962} & 53.19\% & 52.75\% & 53.19\% & 54.33\% & 52.83\% & 44.34\% \\
    MindEye2 & \textbf{0.004} & 0.684 & 53.54\% & 51.41\% & 52.09\% & 52.81\% & 51.53\% & 39.34\% \\
    MindBridge & -0.006 & 0.906 & \textbf{57.49\%} & \textbf{55.69\%} & \textbf{55.69\%} & 57.13\% & 56.69\% & 40.32\% \\
    \bottomrule
  \end{tabular*}
\end{table}

\begin{table}[t]
  \centering
  \caption{Attribute-level evaluation on the simple split.
  Reconstruction scores are class-balanced accuracies, reported
  as mean $\pm$ sample standard deviation across seven participants.
  The chance row gives the analytical expected score under uniform
  selection from the corresponding answer options.
  Differences between model configurations are descriptive.}
  \label{tab:simple_attribute_results}
  \begin{tabular}{lccc}
    \toprule
    Method & Color & Position & Visual identity \\
    \midrule
    LDM
      & $30.13 \pm 5.58\%$
      & $24.07 \pm 5.46\%$
      & $2.36 \pm 1.79\%$ \\
    MindEye1
      & $31.65 \pm 7.83\%$
      & $35.34 \pm 8.56\%$
      & $3.38 \pm 1.96\%$ \\
    MindEye2
      & $32.59 \pm 10.47\%$
      & $29.39 \pm 4.73\%$
      & $2.07 \pm 1.25\%$ \\
    MindBridge
      & $35.55 \pm 4.04\%$
      & $26.99 \pm 6.81\%$
      & $2.70 \pm 1.30\%$ \\
    \midrule
    Chance & $25.76\%$ & $20.00\%$ & $1.05\%$ \\
    \bottomrule
  \end{tabular}
\end{table}

\subsection{Simple split}

Table~\ref{tab_simple_results} reports the results on the simple split. This split is designed as a controlled diagnostic setting rather than an easier version of the complex split. It further supports the conclusion from the complex split that current baselines can recover partial information from AVID, but their success depends strongly on training setting and metric type.

The image-based metrics reveal an important limitation. SSIM is much higher than in the complex split, with two methods above 0.90, whereas PixCorr remains close to zero. 
This discrepancy may partly reflect the sparse structure of the simple stimuli, as large low-texture backgrounds can inflate structural similarity, while pixel correlation remains sensitive to exact foreground position, scale, and contrast.
Thus, high SSIM should not be interpreted as accurate recovery of the specified content, and the near-zero PixCorr again suggests that pixel-level reconstruction is not an appropriate primary objective for auditory-induced VMI.

The description-based metrics provide a more direct diagnostic of attribute recovery. TIFA is higher on the simple split than on the complex split, consistent with the fact that explicitly specified attributes such as identity, color, count, text, and position can be expressed as visual questions. However, the scores remain far from ceiling, indicating that current baselines still do not reliably recover all specified attributes. Together, the simple split complements the complex split by showing that semantic recovery remains incomplete even when the intended visual content is tightly constrained.

Moreover, the controlled design of the simple split enables a finer-grained evaluation of reconstruction quality beyond aggregate image--text scores. Because the descriptions explicitly specify visual attributes, we can assess whether reconstructions preserve individual components of the intended content, including color, spatial position, and visual identity. Table~\ref{tab:simple_attribute_results} reports class-balanced accuracy for these attributes, summarized as the mean and sample standard deviation across seven participants.

The results reveal different patterns of attribute recovery across methods. The analytical chance level for color is $25.76\%$ because in test set some stimuli accept two of the four answer options as correct, increasing the expected class-balanced accuracy under uniform random guessing. MindBridge obtained the highest mean color score, whereas MindEye1 obtained the highest mean position and visual-identity scores. Across methods, mean color and position scores exceeded their analytical chance expectations, although recovery remained incomplete and varied across participants. Visual-identity accuracy ranged from $2.07\%$ to $3.38\%$, compared with a chance expectation of $1.05\%$, indicating particularly limited recovery of the intended identity. These findings illustrate the diagnostic value of the controlled stimuli: aggregate semantic alignment can coexist with failures to reconstruct specific attributes, which become apparent when those attributes are evaluated separately.

\section{Data and Code Availability}

We release the de-identified AVID dataset, an example tutorial dataset, and accompanying evaluation code. The full dataset is hosted at \url{https://doi.org/10.57760/sciencedb.33666

        }, and the code repository at \url{https://github.com/lishurui0612/AVID-benchmark}. A validated Croissant metadata file is provided with the release.

All released imaging data are de-identified before distribution. Additional details on file organization, release terms, and related documentation are provided in Appendix~\ref{app:Data_organization}.

\section{Conclusion and Limitation}

We introduced AVID, a dedicated 5T fMRI dataset for auditory-induced VMI decoding collected under a visual-free imagery paradigm.    Based on AVID, we define a benchmark for this task with fixed splits, predefined neural inputs, and a tailored evaluation protocol, and provide baseline evaluations using representative reconstruction methods.    The resulting reconstructions represent an initial step toward auditory-induced imagery decoding, preserving much of the recognizable semantic content in the strongest cases and showing trends that are broadly consistent with the quantitative metrics.    At the same time, there remains a gap compared with perception-oriented decoding, indicating that future progress will likely require models designed more directly for visual mental imagery.

AVID has several scope boundaries. The current release is limited to Mandarin-speaking participants, and the simple split was collected from a subset of the complex split participants. Because AVID contains human neuroimaging data collected during mental imagery, it is intended for research on auditory-induced visual imagery, neural decoding, and evaluation methodology, with additional limitations and responsible-use considerations provided in Appendix~\ref{app:limitation}.

\begin{ack}
This work was supported by the National Science and Technology Major Project of China (Grant No.~2025ZD0217000 to Y.L.); the National Natural Science Foundation of China (Grant No.~32271099 to X.Z.; Grant No.~32371154 to Y.L.); the Research Center for Brain Cognition and Human Development of Guangdong Province (Grant No.~2024B0303390003 to X.Z.); the Shanghai Rising-Star Program (Grant No.~24QA2705500 to Y.L.); Lin Gang Laboratory (Grant Nos.~LG-GG-202402-06 and LGL-1987-18 to Y.L.); and the State Key Laboratory of Advanced Medical Materials and Devices (Grant No.~YGSKL2026ZZ01 to Y.L.).

The authors acknowledge the Advanced MR Imaging Research Laboratory in the School of Biomedical Engineering at ShanghaiTech University for providing the imaging platform and support for this work.

Competing interests: Yuanning Li is a cofounder of XS Vision, and Shiqi Shen is an intern at XS Vision. The other authors declare no competing interests.
\end{ack}

\bibliographystyle{plainnat}
\bibliography{neurips2026}

\newpage
\appendix

\section{Dataset documentation and release}
\label{app:Data_organization}

\subsection{Dataset contents and structure}

The AVID release follows the Brain Imaging Data Structure (BIDS) format and includes the data and metadata needed to reproduce the neural inputs and evaluation protocol used in the main paper. The released materials include raw fMRI data, preprocessed time-series data, GLMSingle beta estimates, anatomical reconstruction outputs, stimulus metadata, ROI masks, train-test split files, and presentation files containing stimulus order and onset information. The stimulus metadata link each stimulus to its spoken description, audio file, source-image information, task split, and presentation index. The dataset description on Science Data Bank summarizes the released components and access procedure.

\paragraph{Directory structure}
At a high level, the released dataset is organized as follows:
\begin{verbatim}
AVID/
  sub-XX/
    anat/
      sub-XX_run-*_T1w.nii.gz
      sub-XX_run-*_T1w.json
    func/
      session-YY/
        sub-XX_ses-YY_task-*_run-*_AP_bold.nii.gz
        sub-XX_ses-YY_task-*_run-*_AP_bold.json
        sub-XX_ses-YY_task-*_run-*_AP_ref.nii.gz
        sub-XX_ses-YY_task-*_run-*_AP_ref.json
        sub-XX_ses-YY_task-*_run-*_PA_bold.nii.gz
        sub-XX_ses-YY_task-*_run-*_PA_bold.json
        sub-XX_ses-YY_task-*_run-*_PA_ref.nii.gz
        sub-XX_ses-YY_task-*_run-*_PA_ref.json

  source_data/
    used_stimuli.tsv

    experiments_scripts/
      sub-XX_scripts.tar

    stimuli/
      Audio stimuli, source images, and localizer stimuli

  derivatives/
    pp_data/
      sub-XX/
        anat/
          sub-XX_avg_T1w.nii.gz
        func/
          minimally_preprocessed time-series files

    pp_betas_surf/
      sub-XX/
        GLMSingle beta estimates in surface space

    pp_betas_vol/
      sub-XX/
        GLMSingle beta estimates in volume space

    freesurfer_results/
      sub-XX_fsresults.tar

    ROIs/
      sub-XX/
        ROI masks of speech perception and linguistic comprehension

    noise_ceiling/
      surf/sub-XX/
        noise-ceiling estimates in surface space
      vol/sub-XX/
        noise-ceiling estimates in volume space

    dataset_official_split/
      surface/
        sub-XX_complex_train.tar
        sub-XX_complex_val.tar
        sub-XX_simple_train.tar
        sub-XX_simple_val.tar
      volume/
        sub-XX_complex_train.tar
        sub-XX_complex_val.tar
        sub-XX_simple_train.tar
        sub-XX_simple_val.tar

\end{verbatim}

\subsection{De-identification and privacy protection}

All released imaging data are de-identified before distribution. Participant identifiers are replaced with random study codes, and the mapping between study codes and participant identities is not released. Direct identifiers, including participant names, dates of birth, height and weight, scan dates, are removed. Anatomical images are defaced before release.

\subsection{Limitations and responsible use}
\label{app:limitation}

AVID should be interpreted within the assumptions of its acquisition and evaluation protocol. The source image associated with each spoken description is a visual reference used for stimulus construction, model supervision, and evaluation. It is not a unique ground truth image for the participant’s mental imagery. Source image metrics and spoken description metrics should therefore be treated as complementary evaluation proxies rather than direct measurements of an exact imagined image.

The current release is limited in population, language, and sampling scope. All participants were young, healthy Mandarin-speaking adults and were pre-screened for relatively vivid self-reported imagery. This selection may bias the benchmark toward individuals with stronger imagery abilities and limits its representativeness across the broader population. Results should not be assumed to generalize to individuals with lower imagery vividness, other languages, age groups, or clinical populations without further validation. The simple split was collected from a subset of the complex-split participants and contains less scan time, so comparisons between the two splits should be interpreted as diagnostic rather than as matched estimates of task difficulty.

AVID contains de-identified human neuroimaging data collected during a mental imagery task and should be handled as sensitive research data. The dataset is intended for research on auditory induced visual imagery, neural decoding, and evaluation methodology. It is not intended for participant re identification, clinical diagnosis, individual level profiling or decision making, or claims about private mental content beyond the task conditions represented in the dataset.

\subsection{AVID license}

CC BY 4.0

\subsection{Existing assets and licenses}

AVID uses existing assets for stimulus construction, MRI preprocessing, baseline implementation, and evaluation. The complex split is built from COCO-CN captions and MS-COCO source images and annotations, following their original licenses and terms. Mandarin auditory stimuli were synthesized using Google Text-to-Speech under the applicable Google Cloud terms. Localizer sentences were selected from the NLP Chinese Corpus. MRI preprocessing and response estimation use established tools including FSL, FreeSurfer, GLMSingle, MATLAB/Psychtoolbox, and GStreamer, and baseline evaluations use released methods and components from LDM, MindEye, MindEye2, MindBridge, CLIP, SigLIP, AltCLIP, TIFA, Inception, and SSIM under their respective licenses. The AVID license applies only to the newly collected fMRI data, dataset metadata, and materials created by the authors, and does not override the licenses or terms of third-party assets.

\subsection{Participant facing materials}

\paragraph{Pre-scan preparation and safety instructions.}
Before scanning, participants reviewed the study information and provided written informed consent. They were then familiarized with the MRI environment, the response button boxes, and the safety squeeze ball. Participants listened to test audio through the MRI-compatible headphones, and the volume was adjusted to ensure that the stimuli were clearly audible over scanner noise while remaining comfortable. They were instructed to keep their head still during scanning, to use the response buttons for vividness ratings during the main task, and to press the safety squeeze ball if they experienced discomfort or wished to stop the experiment. Standard MRI safety instructions were provided before entering the scanner.

\paragraph{Main imagery task instructions.}
Participants were instructed to imagine only the content described in the audio. If an attribute was not mentioned, they were asked not to add extra semantic content, although surface texture was not constrained. They were instructed to form the image as clearly as possible within the available time. When no spatial position was specified, they were asked to place the imagined content near the center of the scene. Participants kept their eyes closed during imagery trials and were allowed to open their eyes only during non-task periods. They were instructed to keep their head still throughout the scan, to imagine a static image rather than a dynamic event, and to imagine the described visual content rather than recalling a personal experience. Participants were also told that they could stop the experiment at any time and should press the safety squeeze ball in case of discomfort or emergency.

\begin{figure}[t]
    \centering
    \includegraphics[width=0.8\linewidth]{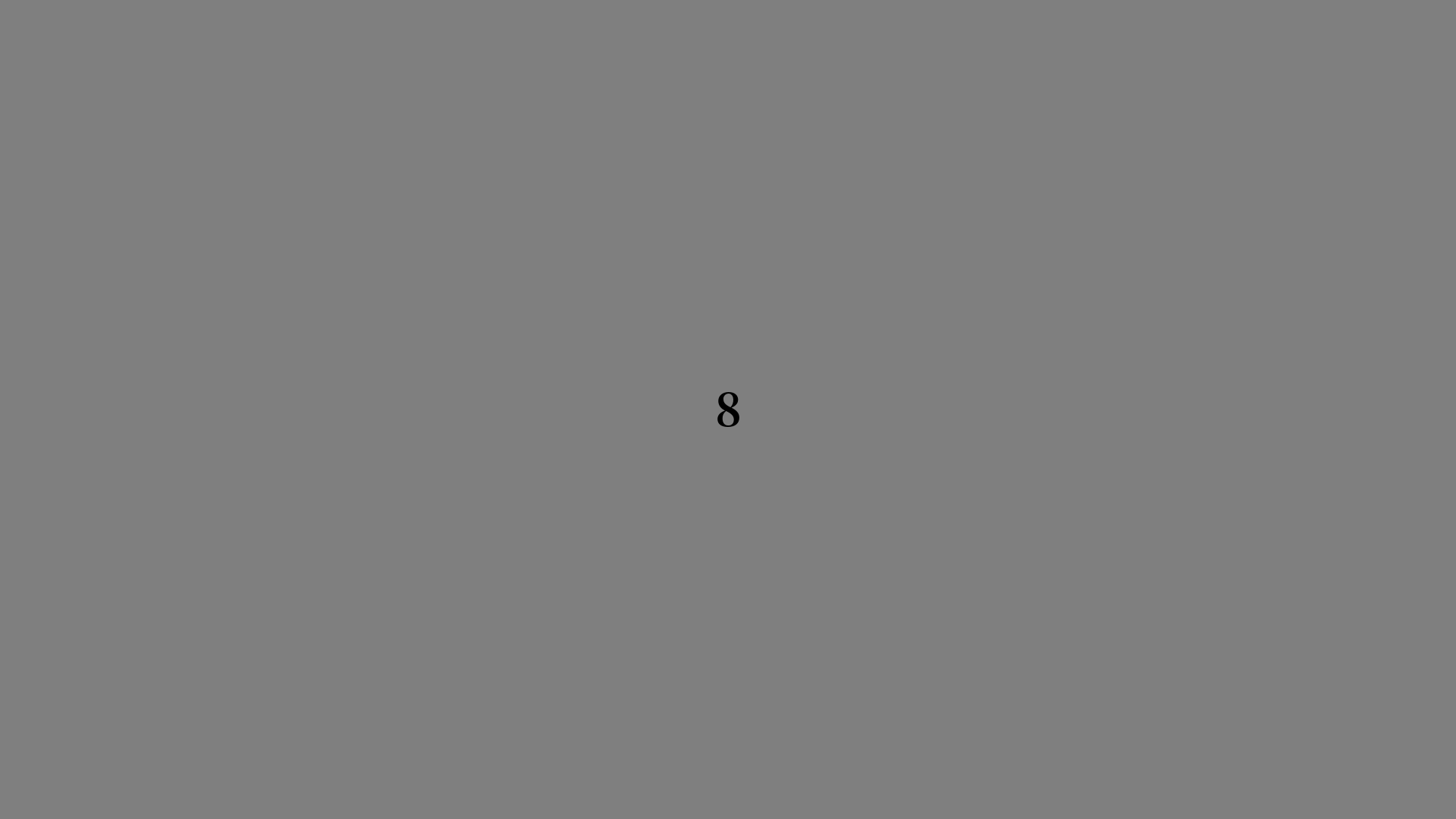}
    \caption{Countdown screenshot.}
    \label{fig:screen}
\end{figure}

Each main imagery session contained eight auditory-imagery runs and one PA reference scan. Each auditory-imagery run lasted approximately 5 minutes. At the beginning of each auditory-imagery run, an 8 s countdown instructed participants to close their eyes before the first auditory prompt, as shown in Figure~\ref{fig:screen}. In each trial, participants listened to a spoken description lasting approximately 4–6 s and formed the corresponding visual image. They then rated the vividness and richness of the imagined image using the response buttons. The rating was completed before the auditory beep, and participants were instructed to maintain the imagined image until the beep occurred. After the beep, participants stopped imagining the previous content and prepared for the next trial after a reset interval. Participants were allowed to take a brief rest every one or two runs.

\paragraph{Functional localizer instructions.}
In the localizer task, participants listened to auditory stimuli while keeping their eyes closed. The stimuli included narrative Mandarin sentences, scrambled Mandarin speech, and natural sounds. For narrative Mandarin sentences, participants were instructed to understand the semantic content. For scrambled Mandarin speech, they were instructed to attend to the pronunciation and acoustic properties without trying to derive semantic meaning. For natural sounds, they were instructed to listen carefully to the sounds. Participants were explicitly instructed not to form visual mental images for any localizer stimuli. In each localizer run, auditory stimuli were presented in repeated blocks. Each block contained four consecutive audio clips, followed by a 14 s interval, and this block structure was repeated until the run ended. Narrative Mandarin and scrambled Mandarin speech were paired within one set of runs, while scrambled Mandarin speech and natural sounds were paired within another set. Each set was repeated twice, resulting in four localizer runs in total. Each localizer run lasted approximately 7 minutes.

\section{Additional experimental results}

\subsection{Stimuli preparation}
\label{app:Stimuli_preparation}
We prepared three sets of auditory stimuli: the complex stimulus split, the simple stimulus split, and the functional localizer stimuli. All task stimuli were presented as spoken Mandarin audio during eyes-closed fMRI scanning.

\paragraph{Complex stimulus split.}
We used the processed image--caption pairs from the Caption Scene Dataset (CSD). From these pairs, we first selected captions containing 12–20 Mandarin characters, resulting in 8,903 candidate pairs. We then manually screened these candidates and retained 1,767 stimulus pairs with clear descriptions and vivid, easily imaginable visual content. The selected captions were converted into audio using Google Text-to-Speech and post-processed to 48 kHz. Finally, 800 stimuli were randomly selected from this curated set. These 800 stimuli were shared across all participants; that is, all participants received the same stimulus set, with only the presentation order varying across participants.

\paragraph{Simple stimulus split.}
The simple split was constructed using rule-based Mandarin caption templates to explicitly control a small set of visual attributes. The actual stimuli presented to participants were all spoken Mandarin descriptions. We used five spatial positions (top, bottom, left, right, and center) and three colors (red, yellow, and blue). The rule-based stimuli included centered uppercase and lowercase letters, centered single digits, centered three-digit number strings, and colored geometric shapes at specified spatial positions. The shape set included triangles, rings, circles, crosses, diamonds, five-pointed stars, hearts, squares, 3-by-3 dot arrays, upward-, downward-, leftward-, and rightward-facing semicircles, vertical rectangles, and horizontal rectangles. This rule-based procedure generated 351 controlled stimuli.

To complement both these rule-based stimuli and the complex scene-level split, we additionally selected 49 high-frequency object categories from ImageNet. These stimuli were designed as single-category imagery trials, in which participants imagined one object category at the center of the scene. Together, the simple split contained 400 stimuli and provided a controlled setting for testing whether reconstructions could recover explicitly specified attributes, including object identity, color, alphanumeric content, shape, count, and position.

\paragraph{English caption preparation.}
All original stimuli in AVID were prepared and presented in Mandarin. Although the source images were derived from COCO-related resources, not every selected image--caption pair retained a directly matched original English annotation. We therefore prepared English captions by translating the Mandarin captions with LLM assistance. Specifically, we used GPT-5.4 to generate structure-preserving English translations. The prompt was:

\begin{tcolorbox}[
    colback=gray!8,
    colframe=gray!40,
    boxrule=0.4pt,
    arc=2pt,
    left=6pt,
    right=6pt,
    top=6pt,
    bottom=6pt
]
\small

\ttfamily
Add a new field named translated\_caption to the TSV file. Translate the Chinese text in the caption field into English, making the translation as fluent and literal as possible. Do not add extra information, and preserve the original structural content.
\end{tcolorbox}

Two authors manually reviewed and cross-checked the English translations against the original Mandarin descriptions and revised them where necessary to preserve semantic accuracy. The translated captions were used as auxiliary annotations for documentation, model input, and text-based evaluation. They were never presented to participants during scanning. For the simple split, this procedure nearly fully preserved the predefined template structure of the original Mandarin captions, including position, color, alphanumeric content, shape identity, and object category (Table~\ref{tab:complex_examples} and~\ref{tab:simple_examples}).

\begin{CJK*}{UTF8}{gbsn}
\begin{table}
\centering
\caption{Example complex stimuli.}
\label{tab:complex_examples}
\begin{tabular}{m{0.18\linewidth} p{0.34\linewidth} p{0.38\linewidth}}
\toprule
Image & Chinese caption & English caption \\
\midrule
\includegraphics[width=\linewidth]{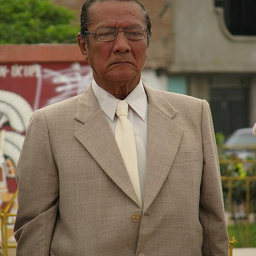}
& 这个男人穿着西装、系着领带、戴着眼镜。
& This man is wearing a suit, a tie, and glasses. \\

\includegraphics[width=\linewidth]{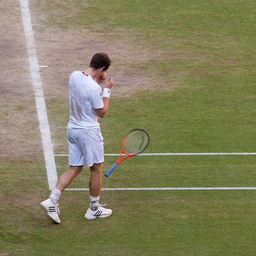}
& 一个网球运动员在草地上丢下了他的球拍
& A tennis player has dropped his racket on the grass. \\

\includegraphics[width=\linewidth]{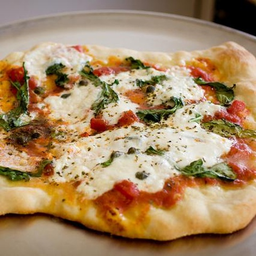}
& 盘子上有一个涂有奶酪的比萨饼
& There is a cheese pizza on the plate. \\

\includegraphics[width=\linewidth]{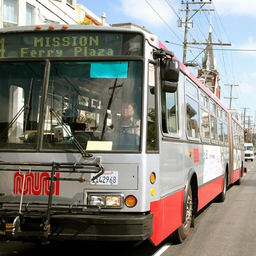}
& 一辆公交车正在道路上行驶。
& A bus is driving on the road. \\

\includegraphics[width=\linewidth]{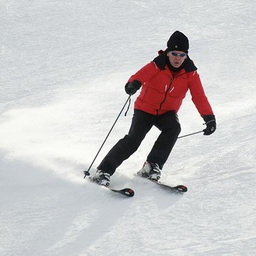}
& 一位男子在雪地里玩滑雪板。
& A man is snowboarding in the snow. \\
\bottomrule
\end{tabular}
\end{table}
\end{CJK*}

\begin{CJK*}{UTF8}{gbsn}
\begin{table}[t]
\centering
\caption{Example simple stimuli.}
\label{tab:simple_examples}
\begin{tabular}{m{0.18\linewidth} p{0.34\linewidth} p{0.38\linewidth}}
\toprule
Image & Chinese caption & English caption \\
\midrule
\includegraphics[width=\linewidth]{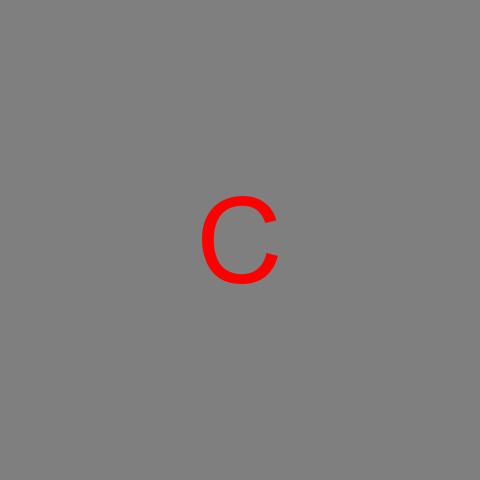}
& 在中间，有一个红色的大写字母C
& In the center, there is a red uppercase letter C. \\

\includegraphics[width=\linewidth]{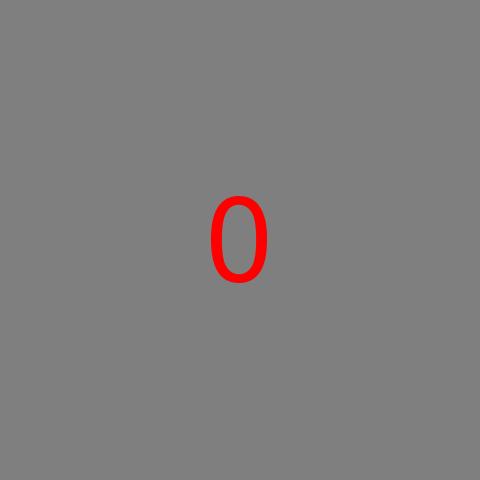}
& 在中间，有一个红色的数字0
& In the center, there is a red digit 0. \\

\includegraphics[width=\linewidth]{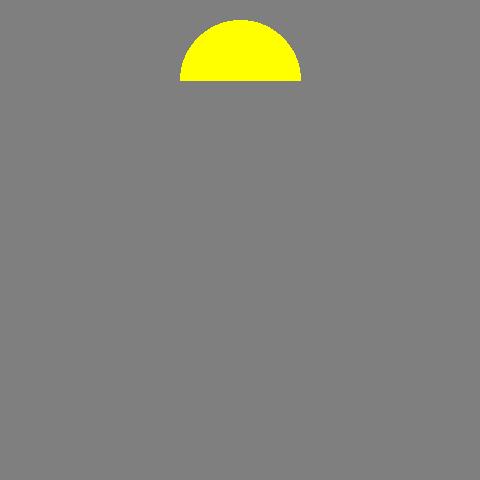}
& 在上方,有一个黄色的朝上的半圆
& At the top, there is a yellow upward-facing semicircle. \\

\includegraphics[width=\linewidth]{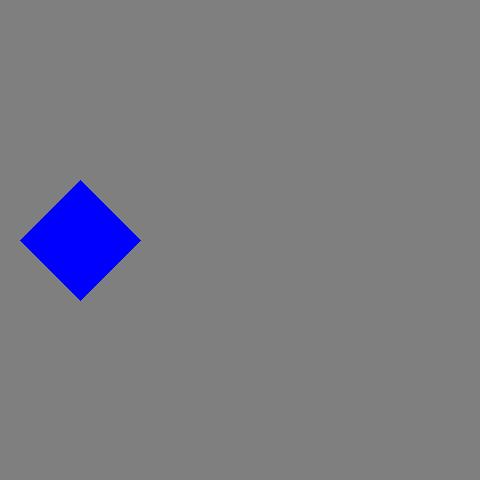}
& 在左边,有一个蓝色的菱形
& On the left, there is a blue diamond. \\

\includegraphics[width=\linewidth]{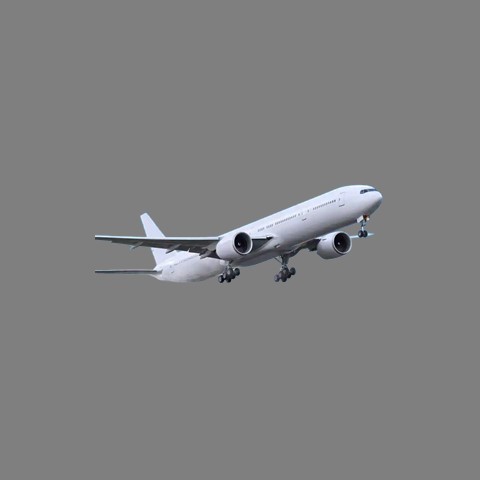}
& 在中间，有一架飞机
& In the center, there is an airplane. \\
\bottomrule
\end{tabular}
\end{table}
\end{CJK*}

\paragraph{Functional localizer stimuli.}
The functional localizer was designed to separate auditory, speech-related, and language-related components elicited by the spoken-description imagery paradigm. It included three auditory conditions: Mandarin sentences, non-word speech-like sequences, and non-linguistic acoustic controls.

For the sentence condition, we selected sixteen Mandarin sentences with low imagery-evoking potential from a Chinese NLP corpus \citep{xu2019nlp}. These sentences were chosen to contain meaningful linguistic content while minimizing explicit visual imagery demands, so that the localizer primarily targeted speech perception and language comprehension rather than visual scene construction.

For the non-word condition, we generated speech-like control sequences from the sentence recordings using an audio-quilting procedure. This procedure rearranged short temporal segments of the original speech recordings to disrupt lexical and semantic content while approximately preserving low-level acoustic properties, including the temporal amplitude envelope and speech-like acoustic structure. The resulting stimuli therefore retained many auditory properties of spoken Mandarin but no longer formed intelligible sentences.

For the non-linguistic acoustic control condition, we used natural sound clips that did not contain speech. Together, these three conditions allowed us to estimate speech-responsive vertices by contrasting speech-like stimuli with non-linguistic sounds, and language-selective vertices by contrasting meaningful Mandarin sentences with non-word speech-like sequences.

\subsection{Experiment procedure}
\label{app:ExperimentProcedure}

\paragraph{Participant screening and pilot experiments.}
Before MRI data collection, participants completed the Vividness of Visual Imagery Questionnaire (VVIQ), as described in Section~\ref{participants}. We selected participants with relatively high self-reported imagery vividness for the MRI experiment. Before finalizing the task protocol, we also conducted pilot experiments using the same auditory-induced imagery paradigm to compare eyes-open fixation and eyes-closed imagery. In the eyes-open variant, participants fixated on a central cross while performing the imagery task. In the eyes-closed variant, participants performed the task with their eyes closed. Based on the pilot results, we adopted the eyes-closed paradigm for the final experiment.

\paragraph{Run and session organization.}
In the final task design, each run contained 25 trials. Each trial lasted 12 s, comprising an 8 s imagery window followed by a 4 s reset interval. Participants rated the vividness of their imagery before the auditory beep at 8 s and maintained the imagined scene until the beep. After the beep, they stopped imagining and rested until the next trial. Each run began with an 8 s blank period before the first trial and ended with a 4 s blank period after the final trial, resulting in a total run duration of 5 min 12 s. Each scanning session consisted of eight task runs. A posterior-to-anterior phase-encoding reference scan was acquired in the middle of each session, after the first four task runs, to support distortion correction.

\paragraph{Scan schedule.}
During the formal MRI experiment, participants completed both structural MRI and task-fMRI scanning. For participants 1--4, structural scans were acquired in separate sessions, interleaved with an unpublished retinotopic localizer that is not included in the AVID release; each of these participants completed four T1-weighted scans. For the remaining participants, two T1-weighted scans were acquired during the first functional MRI session, one at the beginning and one at the end of the session. For the complex split, each participant completed eight task sessions, corresponding to 64 task runs in total. For the simple split, each participant completed four task sessions, corresponding to 32 task runs in total. Participants also completed functional localizer scans, including two runs for the speech perception localizer and two runs for the linguistic comprehension localizer.

\subsection{MRI data acquisition}
\label{app:MRIacquisition}
MRI data were acquired on a 5T United Imaging uMR Jupiter system using a 48-channel head coil. The MRI acquisition included high-resolution T1-weighted anatomical imaging and task-based fMRI. 

During MRI scanning, all participants wore earplugs to reduce scanner noise. During task-fMRI runs, participants additionally wore MRI-compatible headphones to listen to the auditory stimuli. Before scanning, the audio volume was adjusted for each participant to ensure that the stimuli were clearly audible over scanner noise while remaining comfortable. Stimulus presentation was synchronized with scanner triggers to ensure accurate alignment between auditory stimulus onset and fMRI acquisition timing. All auditory stimuli were presented as 48-kHz audio files. Stimulus presentation was implemented in MATLAB using Psychtoolbox 3.0.18 and GStreamer 1.18.6.

Responses were collected using two button boxes, one for each hand, with two buttons on each box. Button responses and response times were recorded throughout the task. A safety squeeze ball was placed next to the participant throughout the experiment, allowing participants to report unexpected discomfort or emergency situations at any time.

Structural T1-weighted MRI images were acquired using a GRE-FSP sequence. The imaging parameters were as follows: voxel size = 0.7 mm isotropic, number of sagittal slices = 270, field of view (FOV) = 221.2 $\times$ 257.6 mm, repetition time (TR) = 7.8 ms, echo time (TE) = 2.6 ms, inversion time (TI) = 950 ms, flip angle = 9 degrees, and bandwidth = 220 Hz per pixel.

Whole-brain functional MRI data were acquired using a T2*-weighted gradient-echo, multi-band accelerated echo-planar imaging (EPI) sequence. The imaging parameters for task-based functional MRI were as follows: number of axial slices = 75, voxel size = 2.0 mm isotropic, matrix size = 100 $\times$ 100, field of view (FOV) = 200 $\times$ 200 mm, slice thickness = 2.0 mm, no slice gap, repetition time (TR) = 1,500 ms, echo time (TE) = 22 ms, flip angle = 62 degrees, echo spacing = 0.335 ms, bandwidth = 1,740 Hz per pixel, multiband factor = 3, and in-plane acceleration factor = 2. Task fMRI runs were acquired with an anterior-to-posterior (AP) phase-encoding direction, and matched posterior-to-anterior (PA) reference images with the same imaging parameters were acquired for distortion correction.

\subsection{tSNR}
\label{app:tSNR}

\begin{figure}[t]
    \centering
    \includegraphics[width=1\linewidth]{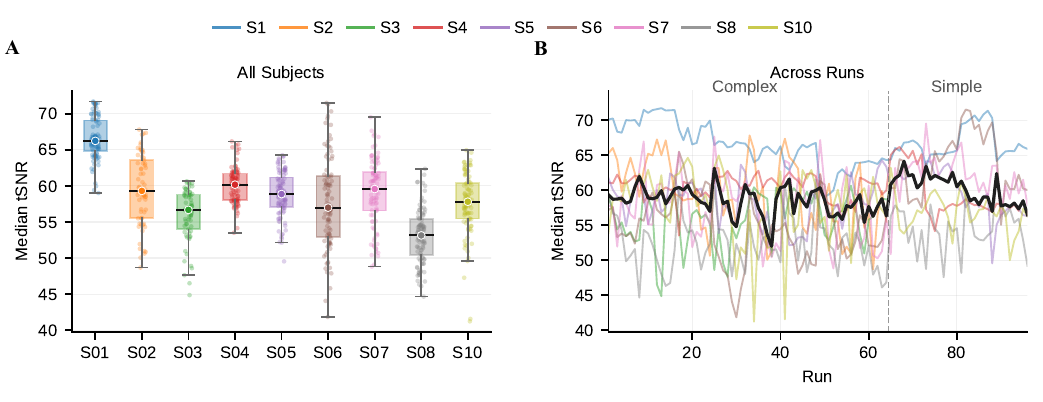}
    \caption{Temporal signal-to-noise ratio (tSNR) across subjects and runs in AVID. (A) Distribution of run-wise median tSNR values for each subject, where tSNR was computed voxel-wise after quadratic detrending and summarized within the top 10\% highest-tSNR voxels in each run. (B) Run-by-run median tSNR trajectories for all subjects, with colored lines indicating individual subjects and the black line indicating the across-subject median. The dashed vertical line separates complex runs from simple runs.}
    \label{fig:tsnr}
\end{figure}
We assessed signal quality by computing the temporal signal-to-noise ratio (tSNR) on the raw fMRI time series from all participants. For each run, tSNR was computed voxel-wise as the temporal mean divided by the temporal standard deviation. To obtain a simple brain mask without relying on preprocessing-derived masks, we first computed the mean functional volume and selected the top 10\% of voxels by mean signal intensity. This threshold was chosen as a coarse high-signal brain mask, considering that the effective cortical voxels occupied approximately 7\% of the full imaging volume.

For each run, we summarized signal quality by taking the median tSNR across voxels within this mask. As shown in Figure~\ref{fig:tsnr}, all participants showed run-wise median tSNR values above 50. Given the relatively short TR and high spatial resolution used in our 5T acquisition, this level of tSNR indicates acceptable raw signal quality. Across runs, tSNR was generally stable for most participants, although participant S6 showed relatively larger run-to-run variability.

\subsection{fMRI preprocessing and data censoring}
\label{app:fMRIpreprocessing}
For participants with multiple T1-weighted scans, individual T1 images were first aligned across scans using FLIRT and then averaged to obtain a high signal-to-noise anatomical reference image. The averaged T1-weighted image was used for cortical surface reconstruction with FreeSurfer's \texttt{recon-all}. The resulting cortical reconstructions were manually inspected and edited when necessary to ensure accurate gray--white matter segmentation. Because the functional images contained 75 slices, one zero-filled slice was appended to each fMRI volume, resulting in 76 slices before distortion correction and subsequent preprocessing.

Functional MRI preprocessing was performed in several steps. First, slice-timing correction was applied using FSL's \texttt{slicetimer}. For each run, the acquired multiband reference image was used as the functional reference image, rather than the temporal mean or median of the fMRI time series. Susceptibility-induced distortion was corrected using FSL's \texttt{topup}, based on the anterior-to-posterior (AP) and posterior-to-anterior (PA) images acquired within the same session. Motion correction was then performed, and all runs from all sessions of the same participant were aligned to a common cross-session functional reference image to improve inter-session stability. The functional reference image was subsequently aligned to the participant's averaged T1-weighted anatomical image using boundary-based registration (BBR). Finally, the estimated transformation matrices from the preprocessing steps were concatenated and applied in a single resampling step to reduce interpolation-related errors. After the minimal preprocessing steps were completed in volume space, the functional signals were projected onto the cortical surface. Following preprocessing, the results of distortion correction and motion correction were manually inspected to ensure data quality. 

Notably, we did not perform volume-level censoring or remove individual fMRI volumes during offline preprocessing. Instead, data quality was monitored during acquisition using the real-time head-motion monitoring system provided by United Imaging. If head motion exceeded the predefined warning threshold during a scan, the corresponding run was excluded from further analysis.

\subsection{Beta estimation}
Single-trial beta estimates were obtained using GLMSingle. Each scanning session consisted of eight runs, and each run contained 25 trials. Each trial lasted 12\,s, with an 8\,s blank period at the beginning of each run and a 4\,s blank period at the end. During stimulus preparation, we followed the sampling strategy used in the NSD, in which stimulus repetitions were generated from a probability distribution formed by mixing a relatively narrow von Mises distribution with a uniform distribution. This design ensured that repeated stimuli were present within each session, which supports the estimation procedure in GLMSingle.

For beta estimation, data from each session were entered into GLMSingle independently. We used the Type-D beta estimates as the released single-trial response estimates in the dataset. The modeled stimulus duration was set to 4\,s. Although the actual auditory stimulus durations were not strictly identical across trials, they were all approximately 4\,s, within a range that can be reasonably modeled by GLMSingle. In the released data, we provide both the original GLMSingle beta estimates and beta estimates that were z-scored within each session.

\subsection{Noise ceiling}
\label{app:nc}
\begin{figure}[t]
    \centering
    \includegraphics[width=\linewidth]{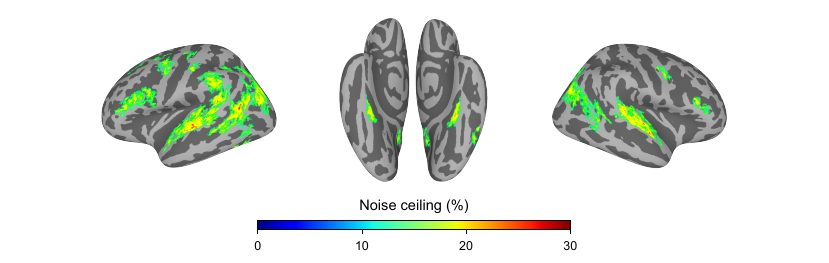}
    \caption{Cortical distribution of noise ceilings, shown in left lateral,
    ventral, and right lateral views. Colors indicate the estimated percentage
    of explainable variance in the mean response across two stimulus repetitions.
    Gray shading depicts cortical anatomy.}
    \label{fig:noise_ceiling}
\end{figure}

We estimated vertex-wise noise ceilings from the two responses to each stimulus. Noise variance was estimated as the mean within-stimulus variance, and signal variance as the total response variance minus noise variance, truncated at zero. The noise ceiling for the two-repeat mean was \[ \mathrm{NC}_v = 100\,\frac{S_v}{S_v + E_v/2}, \] where $S_v$ and $E_v$ denote the estimated signal and noise variances at vertex $v$, respectively.

Individual maps were resampled to fsaverage and averaged across
participants separately for the complex ($n=9$) and simple ($n=7$) splits, excluding the medial wall. Figure~\ref{fig:noise_ceiling} shows the cortical distribution of noise ceilings. With only two repetitions, these estimates provide coarse repeatability diagnostics rather than precise performance upper bounds.

\subsection{Relationship between behavior and neural responses}
\label{app:vividness_relationship}

We examined whether subjective imagery vividness was associated with the reliability of neural responses. Participants reported vividness using a four-point scale during the imagery task. Response frequencies were highly imbalanced across participants, with several participants predominantly selecting a single rating. To avoid sparse estimates from separate rating bins, we ranked stimuli by each participant's mean vividness rating within each split and compared the 100 highest-rated stimuli with the 100 lowest-rated stimuli. We estimated noise ceilings (NCs) separately for these two stimulus groups in four bilateral FreeSurfer-defined cortical regions: the postcentral gyrus, superior temporal gyrus, fusiform gyrus, and pars orbitalis of the inferior frontal gyrus. This analysis included nine participants for the complex split and seven for the simple split. Comparisons between the high- and low-vividness groups used two-sided paired Wilcoxon signed-rank tests across participants.

Table~\ref{tab:vividness_neural_reliability} summarizes the results. In the complex split, the high-vividness group had higher NCs in the postcentral gyrus ($11.55$ versus $8.21$, $p=0.0039$) and superior temporal gyrus ($13.89$ versus $12.38$, $p=0.0273$). The corresponding comparisons in the fusiform gyrus and pars orbitalis yielded $p=0.8203$ and $p=0.3594$, respectively. In the simple split, higher NCs were observed for the high-vividness group in the fusiform gyrus ($9.44$ versus $7.74$, $p=0.0156$) and pars orbitalis ($8.72$ versus $7.15$, $p=0.0156$), whereas the postcentral and superior temporal comparisons yielded $p=0.0781$ and $p=1.0000$.

These results demonstrate a regionally selective association between subjective imagery vividness and neural response reliability in AVID. Higher-vividness stimuli were associated with more reliable responses in the postcentral and superior temporal gyri for the complex split, and in the fusiform gyrus and pars orbitalis for the simple split. This correspondence links the recorded neural responses to participants’ reported imagery experience, with the association expressed in specific cortical regions.

\begin{table}[t]
\centering
\caption{Neural response reliability for the 100 lowest- and 100 highest-vividness stimuli selected separately within each participant and split. Values are across-participant mean $\pm$ sample standard deviation of noise-ceiling estimates. The complex and simple splits included nine and seven participants, respectively. $p$ values were obtained using two-sided paired Wilcoxon signed-rank tests.}
\label{tab:vividness_neural_reliability}
\begin{tabular*}{\linewidth}{@{\extracolsep{\fill}}llccc@{}}
\toprule
Split & Cortical region & Low-vividness NC & High-vividness NC & $p$ \\
\midrule
Complex & Postcentral gyrus       & $8.21 \pm 2.46$\%  & $11.55 \pm 2.64$\% & 0.0039 \\
        & Superior temporal gyrus & $12.38 \pm 1.20$\% & $13.89 \pm 2.16$\% & 0.0273 \\
        & Fusiform gyrus          & $10.57 \pm 3.43$\% & $10.42 \pm 2.84$\% & 0.8203 \\
        & Pars orbitalis          & $8.75 \pm 1.68$\%  & $9.22 \pm 1.88$\%  & 0.3594 \\
\midrule
Simple  & Postcentral gyrus       & $8.94 \pm 2.40$\%  & $9.69 \pm 2.00$\%  & 0.0781 \\
        & Superior temporal gyrus & $12.08 \pm 1.42$\% & $12.06 \pm 2.42$\% & 1.0000 \\
        & Fusiform gyrus          & $7.74 \pm 1.63$\%  & $9.44 \pm 2.22$\%  & 0.0156 \\
        & Pars orbitalis          & $7.15 \pm 1.88$\%  & $8.72 \pm 2.05$\%  & 0.0156 \\
\bottomrule
\end{tabular*}
\end{table}

\subsection{Neural signal validation}
\label{app:neural_validation}

\paragraph{Direct decoding of color and spatial position.}
To assess whether AVID contains decodable neural information without relying on an image-generation model, we decoded color and spatial position directly from surface-space fMRI beta responses in the simple split. Repeated responses to the same stimulus were averaged, and high-reliability vertices were selected using the simple-split noise-ceiling map. Decoding used five-fold stimulus-level grouped cross-validation, so responses to the same stimulus were not shared between training and test folds. For each of the seven participants, we calculated the cross-validated accuracy among the evaluated ridge-classifier and SVM settings. As a random-feature control, we replaced fMRI inputs with Gaussian features while keeping the labels, folds, feature dimensionality, and decoding procedure unchanged.

Table~\ref{tab:direct_attribute_decoding} summarizes the results. Mean decoding accuracy was $0.395 \pm 0.021$ for color and $0.440 \pm 0.144$ for position, compared with nominal chance levels of $1/3$ and $1/5$, respectively. One-sided one-sample $t$-tests against nominal chance yielded $t=7.71$, $p=0.00012$ for color and $t=4.42$, $p=0.0022$ for position; Wilcoxon signed-rank tests yielded $p=0.0078$ for both attributes. The Gaussian-feature controls remained close to nominal chance. 

\begin{table}[t]
\centering
\caption{Direct attribute decoding from simple-split fMRI responses and Gaussian-feature controls. fMRI accuracies are means $\pm$ sample SD across seven participants; Gaussian values summarize the random-feature control. The $p$ values compare participant-level fMRI accuracies with nominal chance and are uncorrected.}
\label{tab:direct_attribute_decoding}
\begin{tabular*}{\linewidth}{@{\extracolsep{\fill}}lccccc@{}}
\toprule
Attribute & fMRI accuracy & Chance & Gaussian control
          & $t$-test $p$ & Wilcoxon $p$ \\
\midrule
Color
  & $0.395 \pm 0.021$ & $0.333$ & $0.332 \pm 0.031$
  & $0.00012$ & $0.0078$ \\
Position
  & $0.440 \pm 0.144$ & $0.200$ & $0.200 \pm 0.031$
  & $0.0022$ & $0.0078$ \\
\bottomrule
\end{tabular*}
\end{table}

\paragraph{fMRI-to-text decoding.} We decoded captions from fMRI responses to 160 held-out complex stimuli per participant across nine participants using a fixed Qwen caption decoder. We evaluated the decoded captions using BLEU-1, BLEU-4, ROUGE-L F1, and Token-F1, commonly used metrics for evaluating reconstructed or generated text against reference captions. BLEU-1 and BLEU-4 measure n-gram precision up to unigrams and four-grams, respectively; ROUGE-L F1 measures longest-common-subsequence overlap; and Token-F1 combines token-level precision and recall. These metrics do not have a universal analytical chance level. We therefore estimated empirical chance performance using within-participant shuffled fMRI and independent Gaussian inputs matched to the training-response mean and standard deviation of each input feature. We generated 50 repetitions of each control per participant while keeping the decoder and generation and evaluation procedures fixed. As shown in Table~\ref{tab:caption_neural_validation}, real fMRI outperformed both controls on all four metrics (all Holm-adjusted $p=0.03125$), supporting a contribution from stimulus-specific neural information beyond the fixed decoder and uninformative inputs.

\begin{table}[t]
\centering
\caption{fMRI-to-text decoding with neural-input controls. Values are means $\pm$ sample SD across nine participants. Bold indicates significantly higher performance than both controls after Holm correction.}
\label{tab:caption_neural_validation}
\begin{tabular*}{\linewidth}{@{\extracolsep{\fill}}lcccc@{}}
\toprule
Input & BLEU-1 & BLEU-4 & ROUGE-L F1 & Token-F1 \\
\midrule
Real fMRI
  & $\mathbf{0.2236 \pm 0.0199}$
  & $\mathbf{0.0759 \pm 0.0079}$
  & $\mathbf{0.2522 \pm 0.0137}$
  & $\mathbf{0.2961 \pm 0.0193}$ \\
Gaussian fMRI
  & $0.2049 \pm 0.0075$
  & $0.0609 \pm 0.0025$
  & $0.2298 \pm 0.0050$
  & $0.2756 \pm 0.0033$ \\
Shuffled fMRI
  & $0.1924 \pm 0.0031$
  & $0.0608 \pm 0.0014$
  & $0.2266 \pm 0.0038$
  & $0.2645 \pm 0.0032$ \\
\bottomrule
\end{tabular*}
\end{table}

\paragraph{Shuffled-fMRI image reconstruction.} We additionally tested whether image reconstruction depended on the correspondence between neural responses and target stimuli. This control was conducted only for MindEye1 on the complex split, across nine participants. For each participant, we generated 50 shuffled-fMRI repetitions by randomly permuting stimulus--response correspondences within that participant while keeping the trained model and reconstruction and evaluation procedures fixed. For each repetition, scores were first averaged across participants; the empirical chance distribution was then summarized across the 50 group-level shuffled scores. Real-fMRI performance was compared with this null distribution using a one-sided empirical permutation test. As shown in Table~\ref{tab:image_shuffle_validation}, the observed group mean exceeded all 50 shuffled group means for every reported metric, with larger differences for semantic identification than for PixCorr and SSIM.

\begin{table}[t]
\centering
\caption{MindEye1 reconstruction with real and shuffled fMRI on the complex split. Real-fMRI values are means $\pm$ sample SD across nine participants; shuffled-fMRI values are means $\pm$ SD across 50 group-level shuffle repetitions. Two-way identification accuracies are reported as percentages. Bold indicates significant real-fMRI performance under a one-sided empirical permutation test.}
\label{tab:image_shuffle_validation}
\begin{tabular}{@{}lcc@{}}
\toprule
Metric & Real fMRI & Shuffled-fMRI chance \\
\midrule
PixCorr
& $\mathbf{0.0562 \pm 0.0245}$
& $0.0272 \pm 0.0072$ \\
SSIM
& $\mathbf{0.2910 \pm 0.0074}$
& $0.2828 \pm 0.0013$ \\
CLIP-I 2WC
& $\mathbf{65.94 \pm 8.20\%}$
& $50.02 \pm 0.91\%$ \\
Inception 2WC
& $\mathbf{65.14 \pm 8.86\%}$
& $49.82 \pm 1.03\%$ \\
CLIP-T 2WC
& $\mathbf{67.83 \pm 8.81\%}$
& $49.90 \pm 0.96\%$ \\
SigLIP 2WC
& $\mathbf{70.63 \pm 10.59\%}$
& $49.99 \pm 0.22\%$ \\
AltCLIP 2WC
& $\mathbf{69.83 \pm 9.59\%}$
& $49.97 \pm 1.02\%$ \\
\bottomrule
\end{tabular}
\end{table}

\subsection{Metric definitions}
\label{app:metrics}

\paragraph{PixCorr.}
PixCorr measures low-level pixel agreement between each reconstruction and its corresponding source image. We compute the Pearson correlation between the flattened pixel values of the reconstructed image and the source image after matching their spatial resolution. The final PixCorr score is averaged over images in the evaluation set. Because auditory-induced imagery does not define a unique pixel-level target, PixCorr is used only as a low-level reference metric rather than as a primary measure of VMI reconstruction quality.

\paragraph{SSIM.}
SSIM measures structural similarity between each reconstruction and its corresponding source image. Unlike PixCorr, SSIM is less sensitive to exact pixel-wise intensity matching and instead captures local luminance, contrast, and structural agreement. The final SSIM score is averaged over images in the evaluation set. As with PixCorr, SSIM is interpreted as a source-image consistency measure, not as direct evidence of the participant's exact imagined image.

\paragraph{Inception 2-Way Comparison (Inception 2WC).}
We report Inception 2WC as a feature-based source-image consistency metric. For each reconstructed image $i$, we compare its Inception feature similarity to the matched source image, $S_{ii}$, against its similarity to each mismatched source image, $S_{ij}$ for all $j \neq i$. The score is the fraction of pairwise comparisons in which the matched source image is more similar:
\[
\mathrm{2WC} = \frac{1}{N(N-1)} \sum_{i=1}^{N} \sum_{j \neq i} \mathbf{1}[S_{ii} > S_{ij}].
\]
This score can be interpreted as an average pairwise two-way identification accuracy, with a chance level of $50\%$.

\paragraph{CLIP-Image 2-Way Comparison (CLIP-I 2WC).}
CLIP-I 2WC uses the same two-way comparison procedure as Inception 2WC, but computes similarities in the CLIP image-embedding space. It evaluates whether each reconstruction is closer to its matched source image than to mismatched source images in a high-level visual feature space. Its chance level is $50\%$.

\paragraph{CLIP-Text 2-Way Comparison (CLIP-T 2WC).}
CLIP-T 2WC evaluates spoken-description alignment using the CLIP image--text embedding space. For each reconstructed image $i$, we compare its similarity to the matched caption, $S_{ii}$, against its similarity to each mismatched caption, $S_{ij}$ for all $j \neq i$, using the same 2WC formula above. The metric measures whether the reconstruction is more aligned with the correct spoken-description caption than with distractor captions. Its chance level is $50\%$.

\paragraph{SigLIP-Text 2-Way Comparison (SigLIP 2WC).}
SigLIP 2WC uses the same image--text two-way comparison procedure as CLIP-T, but computes similarities using SigLIP embeddings. It provides an additional vision-language embedding-based measure of spoken-description alignment. Its chance level is $50\%$.

\paragraph{AltCLIP-Text 2-Way Comparison (AltCLIP 2WC).}
AltCLIP 2WC also follows the same image--text two-way comparison procedure, but uses AltCLIP embeddings to construct the image--text similarity matrix. Because AltCLIP is a multilingual vision-language model, it provides a complementary check for text-based semantic alignment in our Mandarin-stimulus setting. Its chance level is $50\%$.

\paragraph{TIFA.}
We additionally report TIFA as a question-answering-based image--text consistency metric. For each caption, a frozen set of multiple-choice questions is derived from the described content, and a VQA model answers these questions based on the reconstructed image. The per-image TIFA score is computed by the TIFA scorer, and the final reported value is averaged over images in the evaluation set. The number of questions is not fixed across images: in our frozen benchmark, the complex split contains 1,302 questions over 160 images, with an average of 8.14 questions per image and a range of 3 to 18; the simple split contains 418 questions over 80 images, with an average of 5.23 questions per image and a range of 3 to 9. TIFA does not have a single universal chance level because it depends on the number of answer choices per question. In our question banks, all questions are either 2-way or 4-way multiple choice, corresponding to random-guessing baselines of $50\%$ and $25\%$, respectively. Under uniform random guessing over answer choices, the expected average score is approximately $36.7\%$ for the complex split and $35.1\%$ for the simple split. This value is a nominal random-answer baseline, not an empirical chance level of the full TIFA pipeline. Because the actual score is produced by a VQA model conditioned on generated images, the effective empirical baseline can be lower when reconstructions do not contain sufficient visual evidence for the queried content.

\subsection{Evaluation of attribute-level recovery}
\label{app:simple_attribute}

\paragraph{Evaluator and prompting protocol.}
We use a frozen Qwen3-VL-8B model to evaluate foreground color, spatial position, and visual identity in simple-split reconstructions. Each image is paired with three fixed attribute-specific prompts, without access to the target caption or reference answer. The task instructions are:
\begin{itemize}
    \item \textbf{Color:} ``Determine the dominant visible color of the main foreground content in the image.'' Options are \texttt{red}, \texttt{yellow}, \texttt{blue}, and \texttt{other or multicolored}.
    \item \textbf{Position:} ``Determine the location of the main foreground content relative to the entire image canvas.'' Options are \texttt{top}, \texttt{bottom}, \texttt{left}, \texttt{right}, and \texttt{center}.
    \item \textbf{Visual identity:} ``Identify the main foreground content in the image.'' The full prompt includes 95 options covering letters, digit sequences, geometric shapes, and objects, such as \texttt{C}, \texttt{194}, \texttt{triangle}, and \texttt{coffee cup}.
\end{itemize}
Each prompt also provides attribute-specific classification rules and the complete option list. For example, color is judged independently of object identity, and position is determined from the center of the foreground content. All prompts require: ``Output only the exact option text.'' Inference uses greedy decoding (\texttt{do\_sample=False}) with a maximum of 24 generated tokens. After trimming leading and trailing whitespace, responses are scored by exact matching against accepted reference answers. Responses outside the option list are incorrect, without semantic matching or post-hoc correction.

\paragraph{Metric calculation.}
We report class-balanced accuracy separately for color, spatial position, and visual identity:
\[
\mathrm{BA} = \frac{1}{C}\sum_{c=1}^{C}\mathrm{Acc}_c,
\]
where $C$ is the number of reference classes represented in the evaluated images and $\mathrm{Acc}_c$ is the exact-match accuracy within class $c$. Each class therefore receives equal weight regardless of its frequency. Responses must exactly match a predefined accepted answer; invalid responses are scored as incorrect. For 11 natural-object stimuli, both the annotated primary color and \texttt{other or multicolored} are accepted; all other questions have a single accepted answer. Across the full 400-image question bank, uniform random guessing yields class-balanced chance levels of 25.57\%, 20\%, and 1.05\% for color, position, and visual identity, respectively.

\paragraph{Evaluator ceiling and failure cases.}
We evaluated the frozen evaluator on all 400 original reference images using the same prompts and scoring rules. Class-balanced accuracy was 100\% for color and position and 98.95\% for visual identity (96.25\% ordinary accuracy). All 15 identity errors involved semicircle orientation. In 13 cases, the predicted straight-edge direction coincided with the foreground's position on the canvas, suggesting confusion between global position and local orientation. Table~\ref{tab:evaluator_failure_cases} presents representative cases. Each prediction was a valid closed-set response specifying an incorrect straight-edge direction: for example, ``top'' in the table denotes the full response ``semicircle (straight edge at top)''. These results characterize the evaluator's residual errors on reference images and serve as empirical ceilings.

\begin{table}[t]
\centering
\caption{Representative evaluator errors on original reference images. The evaluator correctly identifies color and spatial position but misclassifies semicircle orientation. Identity answers are abbreviated to the straight-edge direction; the full answer format is ``semicircle (straight edge at [direction])''. Reference descriptions are shown for illustration and are not provided to the evaluator.}
\label{tab:evaluator_failure_cases}
\renewcommand{\arraystretch}{1.2}

\begin{tabular}{
@{}
>{\raggedright\arraybackslash}m{0.17\linewidth}
>{\raggedright\arraybackslash}m{0.39\linewidth}
>{\raggedright\arraybackslash}m{0.17\linewidth}
>{\raggedright\arraybackslash}m{0.17\linewidth}
@{}
}
\toprule
Reference image & Reference description & Correct answer & Evaluator answer \\
\midrule

\includegraphics[width=0.9\linewidth]{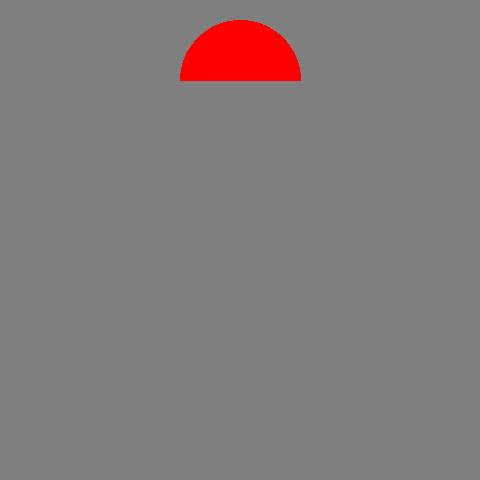}
&
At the top, there is a red upward-facing semicircle.
&
bottom
&
top
\\

\midrule

\includegraphics[width=0.9\linewidth]{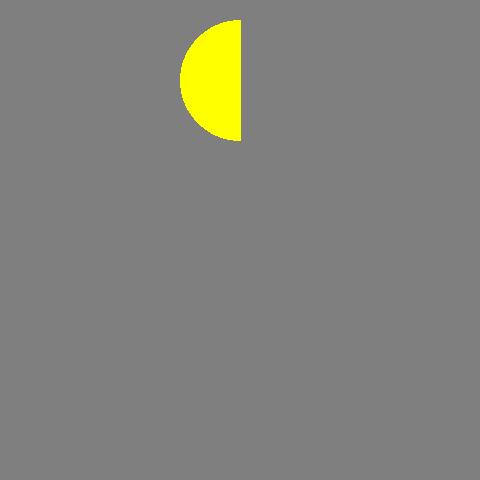}
&
At the top, there is a yellow left-facing semicircle.
&
right
&
top
\\

\midrule

\includegraphics[width=0.9\linewidth]{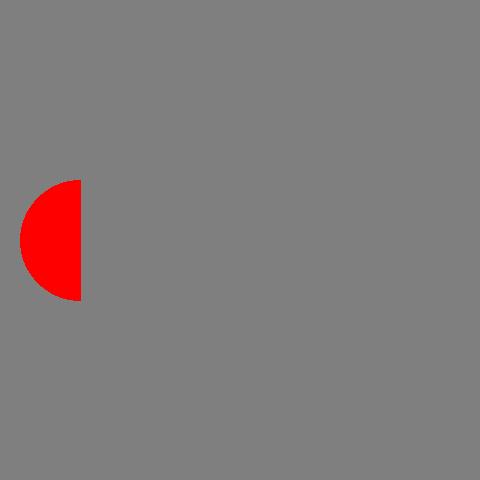}
&
On the left, there is a red left-facing semicircle.
&
right
&
left
\\

\bottomrule
\end{tabular}
\end{table}

\subsection{ROI profiles and ROI-restricted reconstruction}
\label{app:roi_restricted_reconstruction}

To assess whether reconstruction performance could be explained solely by responses in speech- and language-responsive cortex, we compared three functionally defined inputs: the speech ROI, the language ROI, and the remaining cortical ROI outside the union of the speech and language localizer masks. The localizer was conducted under eyes-closed auditory conditions, during which participants were instructed not to form visual images. The remaining ROI therefore represents a speech/language-excluded cortical input rather than an anatomically restricted visual-only ROI.

Vertex selection was performed separately for each participant and task. Within each candidate region, vertices were ranked by noise ceiling, and the top 15\% were retained, with the number of selected vertices rounded upward. Table~\ref{tab:roi_sizes} reports the resulting vertex counts. The speech ROI was concentrated in superior temporal regions, particularly the superior temporal gyrus (STG). The language ROI covered frontal and temporal language regions, including the middle frontal gyrus (MFG), inferior frontal gyrus (IFG), anterior temporal cortex, and posterior temporal cortex. The remaining ROI predominantly covered higher-order visual and associative regions along the dorsal and ventral visual pathways, with comparatively limited involvement of primary visual cortex.

\begin{table}[t]
\centering
\caption{ROI sizes after NC-based vertex selection. Values are the mean $\pm$ sample standard deviation of vertex counts across participants (complex: $n=9$; simple: $n=7$). The top 15\% were selected separately within each candidate region.}
\label{tab:roi_sizes}
\begin{tabular}{lccc}
\toprule
Split & Speech ROI & Language ROI & Remaining ROI \\
\midrule
Complex & $1{,}426 \pm 933$ & $2{,}191 \pm 2{,}153$ & $38{,}538 \pm 5{,}900$ \\
Simple & $1{,}561 \pm 1{,}024$ & $2{,}348 \pm 2{,}460$ & $37{,}537 \pm 6{,}344$ \\
\bottomrule
\end{tabular}
\end{table}

We evaluated ROI-restricted reconstruction using MindEye1 on the complex split. A separate model was trained for each participant and ROI for 240 epochs, with the input dimensionality determined by the selected vertex count. Reconstruction used an image-to-image strength of 0.7 and five candidates per stimulus. Performance was evaluated on the 160 held-out stimuli for each of the nine participants and summarized as the participant-level mean $\pm$ sample standard deviation. For each of the ten metrics, we tested whether the remaining ROI outperformed the speech ROI and the language ROI using one-sided paired $t$-tests across participants. Benjamini--Hochberg correction was applied jointly across all 20 comparisons, with significance defined as adjusted $q<0.05$.

\begin{table}[t]
\centering
\caption{ROI-restricted MindEye1 reconstruction performance on the complex split ($n=9$). Values are the mean $\pm$ sample standard deviation across participants. The final two columns report Benjamini--Hochberg-adjusted $p$-values from one-sided paired $t$-tests assessing whether the remaining ROI outperformed the respective comparison ROI. Correction was applied jointly across all 20 comparisons; adjusted $p<0.05$ indicates significant superiority.}
\label{tab:roi_reconstruction}
\begin{tabular*}{\linewidth}{@{\extracolsep{\fill}}lccccc@{}}
\toprule
Metric & Speech ROI & Language ROI & Remaining ROI & vs.\ Speech & vs.\ Language \\
\midrule
PixCorr & $0.046 \pm 0.019$ & $0.049 \pm 0.026$ & $0.061 \pm 0.024$ & $0.0747$ & $0.0915$ \\
SSIM & $0.327 \pm 0.028$ & $0.296 \pm 0.011$ & $0.291 \pm 0.011$ & $0.9976$ & $0.8892$ \\
CLIP-I 2WC & $0.538 \pm 0.064$ & $0.613 \pm 0.069$ & $0.657 \pm 0.083$ & $0.0011$ & $0.0136$ \\
Inception 2WC & $0.537 \pm 0.062$ & $0.602 \pm 0.081$ & $0.648 \pm 0.092$ & $0.0030$ & $0.0257$ \\
CLIP-T R@1 & $0.013 \pm 0.013$ & $0.026 \pm 0.032$ & $0.035 \pm 0.039$ & $0.0488$ & $0.1294$ \\
CLIP-T R@5 & $0.057 \pm 0.055$ & $0.092 \pm 0.083$ & $0.147 \pm 0.105$ & $0.0056$ & $0.0070$ \\
CLIP-T 2WC & $0.554 \pm 0.071$ & $0.612 \pm 0.090$ & $0.681 \pm 0.094$ & $0.0006$ & $0.0039$ \\
SigLIP-T 2WC & $0.558 \pm 0.088$ & $0.633 \pm 0.087$ & $0.700 \pm 0.095$ & $0.0006$ & $0.0030$ \\
AltCLIP-T 2WC & $0.563 \pm 0.079$ & $0.625 \pm 0.094$ & $0.703 \pm 0.096$ & $0.0006$ & $0.0026$ \\
TIFA & $0.254 \pm 0.040$ & $0.277 \pm 0.043$ & $0.309 \pm 0.054$ & $0.0010$ & $0.0035$ \\
\bottomrule
\end{tabular*}
\end{table}

As shown in Table~\ref{tab:roi_reconstruction}, the remaining ROI significantly outperformed both comparison ROIs on seven of the ten metrics: CLIP-I 2WC, Inception 2WC, CLIP-T R@5, CLIP-T 2WC, SigLIP-T 2WC, AltCLIP-T 2WC, and TIFA. For CLIP-T R@1, it significantly outperformed the speech ROI but not the language ROI. Neither comparison established a significant advantage for the remaining ROI on PixCorr or SSIM. These findings indicate that information supporting reconstruction was retained outside the regions identified by the no-imagery speech and language localizer, particularly for high-level visual and semantic measures.
However, differences in vertex count cannot be entirely ruled out as a contributing factor to the remaining ROI's performance advantage. Although this ROI includes higher-order visual and associative regions, the present analysis cannot determine which subregions contribute most to reconstruction.  Future work could use parcel-level decoding and region-wise ablation to quantify their contributions.

\subsection{ROI sensitivity analysis}
\label{app:sensitivity analysis}

We further evaluate the stricter neural input $\mathrm{ROI}_m \setminus (\mathrm{ROI}_s \cup \mathrm{ROI}_l)$ as a sensitivity analysis. This input removes both speech-responsive and language-selective vertices from the task-responsive mask. The goal is not to define a purer VMI region, since language-selective cortex may still carry conceptual information used to construct the imagined scene. Instead, this analysis tests whether the main trends remain stable when auditory and language-responsive components are more aggressively excluded.

\begin{figure}[t]
    \centering
    \includegraphics[width=1\linewidth]{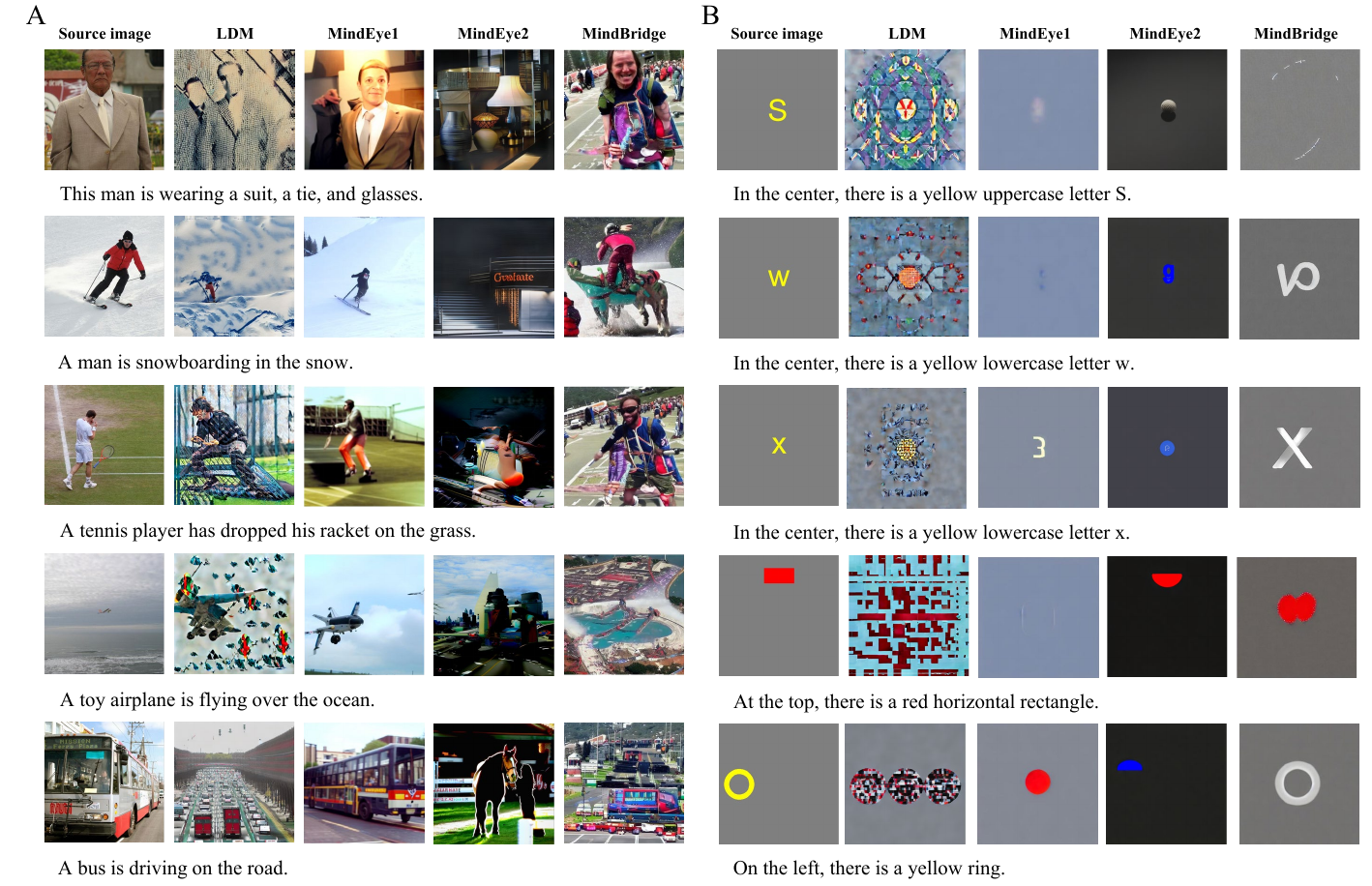}
    \caption{Qualitative comparison of mental imagery generation results for ROI sensitivity analysis. (A) Examples from the complex stimuli split. (B) Examples from the simple stimuli split. For each stimulus, we compare the source image with reconstructions generated by LDM, MindEye1, MindEye2, and MindBridge. The results illustrate how different methods recover semantic content and visual structure across natural images and simple shape stimuli.}
    \label{fig:ReconstructionFigure_nospelan}
\end{figure}

Figure~\ref{fig:ReconstructionFigure_nospelan} shows qualitative reconstructions under the stricter input. Compared with the primary setting, the complex examples show similar qualitative behavior. The degree of semantic recovery is broadly comparable, with several reconstructions still preserving recognizable scene content while remaining variable in layout and local detail. The simple examples show more visible loss of some specified attributes in the selected cases, although these examples should be interpreted together with the aggregate metrics rather than as a standalone ranking. LDM remains qualitatively similar across the two ROI settings, partly because its outputs often contain strong texture and predictions at multiple plausible locations. MindBridge more often preserves semantic identity, but color recovery appears less stable than in some other methods. MindEye2 shows weaker color recovery in these examples, while retaining position information relatively well.

Table~\ref{tab:roi_sensitivity_complex} reports the complex split results. The stricter input produces only modest changes for LDM and MindEye1 across most metrics. 
MindEye1 continues to achieve relatively strong performance across Inception, CLIP-I, CLIP-T, SigLIP, and AltCLIP under the stricter ROI definition.
MindEye2 shows mixed changes across metrics, with no consistent directional shift relative to the primary input. MindBridge shows modest decreases across feature- and text-based identification metrics under the stricter input, while retaining substantial correspondence with the source images and spoken descriptions. This suggests that the multi-participant model is sensitive to the ROI definition, but can still recover decodable information after language-selective vertices are removed.

\begin{table}[t]
  \caption{Complex split baseline results under NSL. Metrics are grouped by source image consistency and spoken description alignment.}
  \label{tab:roi_sensitivity_complex}
  \centering
  \begin{tabular*}{\linewidth}{@{\extracolsep{\fill}}lcccccccc@{}}
    \toprule
    & \multicolumn{4}{c}{Image-image} & \multicolumn{4}{c}{Image-text} \\
    \cmidrule(lr){2-5} \cmidrule(lr){6-9}
    Method & PixCorr & SSIM & Inception & CLIP-I & CLIP-T & SigLIP & AltCLIP & TIFA \\
    \midrule
    LDM & 0.040 & 0.187 & 58.30\% & 59.36\% & 67.42\% & 68.30\% & 68.09\% & 31.99\% \\
    MindEye1 & \textbf{0.061} & \textbf{0.291} & \textbf{64.85\%} & \textbf{65.71\%} & \textbf{68.06\%} & \textbf{70.04\%} & \textbf{70.33\%} & 30.91\% \\
    MindEye2 & 0.031 & 0.265 & 55.06\% & 56.75\% & 55.77\% & 57.85\% & 57.06\% & 24.62\% \\
    MindBridge & 0.040 & 0.170 & 63.69\% & 64.78\% & 66.70\% & 67.82\% & 68.50\% & \textbf{32.06\%} \\
    \bottomrule
  \end{tabular*}
\end{table}

Table~\ref{tab:roi_sensitivity_simple} reports the simple split results. PixCorr remains near zero across methods, and SSIM remains high for MindEye1 and MindBridge. The text-based metrics also show method-specific changes. LDM and MindEye1 remain close to their primary-input results, while MindEye2 shows mixed changes across metrics, with no consistent directional shift. MindBridge retains the strongest CLIP-I score, but its AltCLIP and TIFA scores remain lower than those of the strongest simple-split baselines. These results are consistent with the main analysis in showing that the simple split is not uniformly easier than the complex split, and that recovering explicitly specified attributes remains difficult under the stricter ROI definition.

\begin{table}[t]
  \caption{Simple split baseline results under NSL. Metrics are grouped by source image consistency and spoken description alignment.}
  \label{tab:roi_sensitivity_simple}
  \centering
  \begin{tabular*}{\linewidth}{@{\extracolsep{\fill}}lcccccccc@{}}
    \toprule
    & \multicolumn{4}{c}{Image-image} & \multicolumn{4}{c}{Image-text} \\
    \cmidrule(lr){2-5} \cmidrule(lr){6-9}
    Method & PixCorr & SSIM & Inception & CLIP-I & CLIP-T & SigLIP & AltCLIP & TIFA \\
    \midrule
    LDM & -0.006 & 0.636 & 51.90\% & 52.31\% & \textbf{55.69\%} & 57.12\% & \textbf{56.87\%} & \textbf{48.42\%} \\
    MindEye1 & -0.031 & \textbf{0.963} & 52.35\% & 51.10\% & 51.45\% & 51.15\% & 51.84\% & 44.82\% \\
    MindEye2 & \textbf{0.003} & 0.679 & 54.39\% & 53.93\% & 52.04\% & 53.07\% & 52.85\% & 41.15\% \\
    MindBridge & -0.008 & 0.916 & \textbf{56.84\%} & \textbf{55.53\%} & 54.72\% & \textbf{57.58\%} & 54.59\% & 40.61\% \\
    \bottomrule
  \end{tabular*}
\end{table}

Overall, the stricter ROI analysis supports the main findings. The baselines do not rely only on vertices identified as speech-responsive or language-selective, since recognizable visual and semantic information remains recoverable after these vertices are removed. At the same time, the method-specific changes show that neural input definition is an important part of the evaluation setting. We therefore report both ROI settings to characterize how the baseline evaluations vary with the neural input definition, rather than treating either setting as a definitive isolation of VMI-related activity.

\subsection{Participant-level variability}
\label{app:participant_variability}
Tables~\ref{tab:variability_image} and~\ref{tab:variability_text} report participant-level means and sample standard deviations across both stimulus splits and neural input settings. Each participant contributes one score per metric, and the reported standard deviations quantify between-participant variability.

\begin{table}[t]
\centering
\caption{Participant-level variability in source-image consistency metrics. Values are mean $\pm$ sample SD across nine participants for the complex split and seven for the simple split. Inception and CLIP-I scores are reported as percentages. Bold marks the highest mean within each split and input setting. NS denotes $\mathrm{ROI}_m\setminus\mathrm{ROI}_s$, and NSL denotes $\mathrm{ROI}_m\setminus(\mathrm{ROI}_s\cup\mathrm{ROI}_l)$.}
\label{tab:variability_image}
\renewcommand{\arraystretch}{1.12}
\begin{tabular*}{\textwidth}{@{\extracolsep{\fill}}lcccc@{}}
\toprule
Method & PixCorr & SSIM & Inception & CLIP-I \\
\midrule
\multicolumn{5}{l}{\textbf{Complex split: NS input}} \\
LDM
& $0.038\pm0.014$
& $0.185\pm0.005$
& $59.10\pm5.00\%$
& $59.60\pm4.90\%$ \\
MindEye1
& $\boldsymbol{0.056\pm0.025}$
& $\boldsymbol{0.291\pm0.007}$
& $65.10\pm8.90\%$
& $\boldsymbol{65.90\pm8.20\%}$ \\
MindEye2
& $0.034\pm0.017$
& $0.262\pm0.018$
& $55.70\pm4.70\%$
& $57.80\pm4.30\%$ \\
MindBridge
& $0.040\pm0.006$
& $0.170\pm0.006$
& $\boldsymbol{65.20\pm6.50\%}$
& $65.80\pm7.00\%$ \\
\midrule
\multicolumn{5}{l}{\textbf{Simple split: NS input}} \\
LDM
& $-0.001\pm0.012$
& $0.636\pm0.017$
& $54.10\pm3.20\%$
& $51.70\pm1.40\%$ \\
MindEye1
& $-0.028\pm0.018$
& $\boldsymbol{0.962\pm0.003}$
& $53.20\pm3.40\%$
& $52.80\pm3.20\%$ \\
MindEye2
& $\boldsymbol{0.004\pm0.030}$
& $0.684\pm0.029$
& $53.50\pm3.40\%$
& $51.40\pm1.40\%$ \\
MindBridge
& $-0.006\pm0.014$
& $0.906\pm0.006$
& $\boldsymbol{57.50\pm3.70\%}$
& $\boldsymbol{55.70\pm3.60\%}$ \\
\midrule
\multicolumn{5}{l}{\textbf{Complex split: NSL input}} \\
LDM
& $0.040\pm0.013$
& $0.187\pm0.005$
& $58.30\pm5.70\%$
& $59.40\pm5.00\%$ \\
MindEye1
& $\boldsymbol{0.061\pm0.024}$
& $\boldsymbol{0.291\pm0.011}$
& $\boldsymbol{64.80\pm9.20\%}$
& $\boldsymbol{65.70\pm8.30\%}$ \\
MindEye2
& $0.031\pm0.021$
& $0.265\pm0.014$
& $55.10\pm4.80\%$
& $56.80\pm4.40\%$ \\
MindBridge
& $0.040\pm0.007$
& $0.170\pm0.005$
& $63.70\pm6.50\%$
& $64.80\pm6.80\%$ \\
\midrule
\multicolumn{5}{l}{\textbf{Simple split: NSL input}} \\
LDM
& $-0.006\pm0.011$
& $0.636\pm0.028$
& $51.90\pm3.40\%$
& $52.30\pm0.70\%$ \\
MindEye1
& $-0.031\pm0.016$
& $\boldsymbol{0.963\pm0.003}$
& $52.30\pm1.40\%$
& $51.10\pm1.80\%$ \\
MindEye2
& $\boldsymbol{0.003\pm0.025}$
& $0.679\pm0.023$
& $54.40\pm2.10\%$
& $53.90\pm3.90\%$ \\
MindBridge
& $-0.008\pm0.014$
& $0.916\pm0.005$
& $\boldsymbol{56.80\pm3.60\%}$
& $\boldsymbol{55.50\pm2.60\%}$ \\
\bottomrule
\end{tabular*}
\end{table}

\begin{table}[t]
\centering
\caption{Participant-level variability in spoken-description alignment metrics. All metrics are reported as percentages, with values given as mean $\pm$ sample SD across nine participants for the complex split and seven for the simple split. Bold marks the highest mean within each split and input setting. NS and NSL follow Table~\ref{tab:variability_image}.}
\label{tab:variability_text}
\renewcommand{\arraystretch}{1.12}
\begin{tabular*}{\textwidth}{@{\extracolsep{\fill}}lcccc@{}}
\toprule
Method & CLIP-T & SigLIP & AltCLIP & TIFA \\
\midrule
\multicolumn{5}{l}{\textbf{Complex split: NS input}} \\
LDM
& $66.80\pm8.30\%$
& $68.50\pm7.60\%$
& $67.30\pm8.40\%$
& $\boldsymbol{32.60\pm4.40\%}$ \\
MindEye1
& $\boldsymbol{67.80\pm8.80\%}$
& $\boldsymbol{70.63\pm10.59\%}$
& $\boldsymbol{69.80\pm9.60\%}$
& $30.30\pm4.90\%$ \\
MindEye2
& $57.90\pm5.30\%$
& $59.10\pm5.90\%$
& $58.80\pm6.50\%$
& $24.30\pm1.90\%$ \\
MindBridge
& $67.70\pm7.00\%$
& $69.20\pm7.20\%$
& $69.60\pm7.50\%$
& $32.30\pm3.30\%$ \\
\midrule
\multicolumn{5}{l}{\textbf{Simple split: NS input}} \\
LDM
& $55.60\pm3.80\%$
& $\boldsymbol{57.20\pm1.70\%}$
& $\boldsymbol{57.20\pm3.80\%}$
& $\boldsymbol{48.90\pm2.00\%}$ \\
MindEye1
& $53.20\pm4.20\%$
& $54.30\pm5.20\%$
& $52.80\pm4.10\%$
& $44.30\pm1.50\%$ \\
MindEye2
& $52.10\pm2.50\%$
& $52.80\pm2.60\%$
& $51.50\pm2.10\%$
& $39.30\pm2.70\%$ \\
MindBridge
& $\boldsymbol{55.70\pm4.00\%}$
& $57.10\pm4.20\%$
& $56.70\pm2.60\%$
& $40.30\pm1.50\%$ \\
\midrule
\multicolumn{5}{l}{\textbf{Complex split: NSL input}} \\
LDM
& $67.40\pm8.40\%$
& $68.30\pm8.40\%$
& $68.10\pm8.40\%$
& $32.00\pm3.60\%$ \\
MindEye1
& $\boldsymbol{68.10\pm9.40\%}$
& $\boldsymbol{70.00\pm9.50\%}$
& $\boldsymbol{70.30\pm9.60\%}$
& $30.90\pm5.40\%$ \\
MindEye2
& $55.80\pm4.60\%$
& $57.80\pm6.50\%$
& $57.10\pm5.70\%$
& $24.60\pm2.50\%$ \\
MindBridge
& $66.70\pm8.30\%$
& $67.80\pm8.60\%$
& $68.50\pm8.80\%$
& $\boldsymbol{32.10\pm3.40\%}$ \\
\midrule
\multicolumn{5}{l}{\textbf{Simple split: NSL input}} \\
LDM
& $\boldsymbol{55.70\pm3.50\%}$
& $57.10\pm3.00\%$
& $\boldsymbol{56.90\pm3.40\%}$
& $\boldsymbol{48.40\pm2.20\%}$ \\
MindEye1
& $51.40\pm1.90\%$
& $51.20\pm3.10\%$
& $51.80\pm3.80\%$
& $44.80\pm1.60\%$ \\
MindEye2
& $52.00\pm3.20\%$
& $53.10\pm3.70\%$
& $52.80\pm2.30\%$
& $41.10\pm1.40\%$ \\
MindBridge
& $54.70\pm3.20\%$
& $\boldsymbol{57.60\pm3.60\%}$
& $54.60\pm2.50\%$
& $40.60\pm2.40\%$ \\
\bottomrule
\end{tabular*}
\end{table}

\section{Implementation details}
\label{app:baseline_implementation}

\paragraph{LDM.}
For LDM, we used a \emph{single-subject} training setup. Unlike the neural network baselines, LDM does not use epoch-based stochastic optimization. Instead, it fits linear ridge-regression models from fMRI responses to text embeddings and VAE latents using the \texttt{himalaya} kernel ridge implementation. The default configuration used 5-fold cross-validation over a predefined alpha grid, with \texttt{fit\_intercept=false}. Accordingly, the notions of optimizer, learning rate, weight decay, and number of training epochs are not applicable in the usual deep-learning sense. The default batch size for feature extraction was 4, while inference used Stable Diffusion with 50 denoising steps, guidance scale 7.5, and img2img strength 0.8. Experiments were run on NVIDIA A100 80GB GPUs. LDM used one GPU and required approximately 15 minutes per participant.

\paragraph{MindEye1.}
For MindEye1, we used a \emph{single-subject} training setup. The model was trained for 240 epochs with a batch size of 64 and a maximum learning rate of $3\times10^{-4}$. Optimization was performed with AdamW. Following the original implementation, weight decay was set to $10^{-2}$ for regular parameters and to $0$ for bias and normalization parameters. We used a cycle learning-rate schedule by default. MixCo-based voxel mixup was enabled during the early phase of training with \texttt{mixup\_pct}=0.33, i.e., approximately the first 33\% of epochs, after which training switched to the corresponding soft contrastive objective without MixCo. The unified configuration further enabled image augmentation and used CLIP ViT-L/14 as the target embedding space. Experiments were run on NVIDIA A100 80GB GPUs. MindEye1 used one GPU and required approximately 1.5 hours per participant.

\paragraph{MindEye2.}
For MindEye2, we used the \emph{multi-subject leave-one-out (LOO)} setting as the default benchmark configuration. The main training stage used all non-held-out subjects jointly and was run for 150 epochs with batch size 64 and maximum learning rate $10^{-3}$. Optimization was performed with AdamW with weight decay $10^{-2}$, together with a cycle learning-rate schedule by default. MixCo was used in the main training stage with \texttt{mixup\_pct}=0.33, meaning that voxel-space MixCo augmentation was active for approximately the first 33\% of epochs. After the main LOO training stage, the held-out subject was further fine-tuned for 15 epochs with learning rate $3\times10^{-4}$, no learning-rate schedule, and MixCo disabled (\texttt{loo\_finetune\_mixup\_pct}=0.0). The default MindEye2 configuration used a 4-block backbone with hidden dimension 4096, OpenCLIP token targets, and both the diffusion prior and blurry reconstruction branches. Experiments were run on NVIDIA A100 80GB GPUs. MindEye2 used 16 GPUs and required approximately 1 hour per held-out participant, including the LOO training and fine-tuning stages.

\paragraph{MindBridge.}
For MindBridge, we used the \emph{multi-subject joint} training setup. The default configuration trained the model for 600 epochs with batch size 4 and validation batch size 4, using a maximum learning rate of $1.5\times10^{-4}$. Optimization was performed with AdamW. As in the official implementation, weight decay was set to $10^{-2}$ for regular parameters and to $0$ for bias and normalization parameters. A cycle learning-rate schedule was used by default, with evaluation and checkpointing performed every 10 epochs. In contrast to MindEye1 and MindEye2, MixCo was not used in the default MindBridge setup. The model used a 4-block architecture with hidden size 2048 and incorporated the standard joint reconstruction objective together with the cycle-consistency term from the MindBridge framework. Experiments were run on NVIDIA A100 80GB GPUs. MindBridge used one GPU and required approximately 4 hours per participant.

\paragraph{Baseline summary.}
Table~\ref{tab:baseline_summary} summarizes the training settings, primary neural inputs, and implementation procedures of the four baselines. All methods used the same released AVID stimulus-level splits and evaluation metrics.

\begin{table}[t]
\centering
\caption{Summary of baseline configurations. The primary input ROI is $\mathrm{ROI}_m \setminus \mathrm{ROI}_s$ for all four methods. Detailed hyperparameters are provided in the model-specific descriptions above; the additional speech/language-excluded input is evaluated in the ROI sensitivity analysis.}
\label{tab:baseline_summary}
\begin{tabular*}{\linewidth}{@{\extracolsep{\fill}}p{0.12\linewidth}p{0.19\linewidth}p{0.16\linewidth}p{0.46\linewidth}@{}}
\toprule
Baseline & Training setting & Primary input ROI & Model fitting and reconstruction \\
\midrule
LDM & Single-subject & $\mathrm{ROI}_m \setminus \mathrm{ROI}_s$ & Ridge regression with 5-fold cross-validation over the regularization grid; Stable Diffusion reconstruction with 50 denoising steps, guidance scale 7.5, and img2img strength 0.8. \\
\addlinespace
MindEye1 & Single-subject & $\mathrm{ROI}_m \setminus \mathrm{ROI}_s$ & Subject-specific training for 240 epochs with CLIP ViT-L/14 targets; MindEye1 reconstruction and candidate-selection pipeline. \\
\addlinespace
MindEye2 & Leave-one-participant-out transfer & $\mathrm{ROI}_m \setminus \mathrm{ROI}_s$ & Joint training on non-target participants for 150 epochs, followed by 15 epochs of target-participant fine-tuning; diffusion-prior and blurry reconstruction branches. \\
\addlinespace
MindBridge & Multi-participant joint training & $\mathrm{ROI}_m \setminus \mathrm{ROI}_s$ & Joint training for 600 epochs with reconstruction and cycle-consistency objectives. \\
\bottomrule
\end{tabular*}
\end{table}

\begin{table}[t]
\centering
\renewcommand{\arraystretch}{1.15}
\caption{Identity membership of the additional visual-identity-disjoint split. Counts in parentheses indicate the number of identities. Semicircles are described by the location of their straight edge.}
\label{tab:identity_disjoint_membership}
\begin{tabular}{@{}p{0.13\linewidth} p{0.53\linewidth} p{0.28\linewidth}@{}}
\toprule
Family & Training identities & Validation identities \\
\midrule
Letter & F, G, H, J, K, Q, W, X, Y (9) & C, S (2) \\
\addlinespace
Single digit & 0, 1, 2, 3, 4, 6, 7, 9 (8) & 5, 8 (2) \\
\addlinespace
Digits & 103, 194, 372, 431, 566, 653, 740, 918 (8) & 286, 829 (2) \\
\addlinespace
Shape & Circle; cross; diamond; five-pointed star; ring; square; three-by-three dot array; triangle; vertical rectangle; semicircles with the straight edge at the bottom, left, or right (12) & Heart; horizontal rectangle; semicircle with the straight edge at the top (3) \\
\addlinespace
Object & T-shirt; backpack; bicycle; book; bookcase; bottle; bottle cap; bow; bus; car; cat; chair; coffee cup; desk lamp; dog; duck; electric fan; forklift; fountain; hair dryer; hammer; laptop; microwave; mirror; mobile phone; motorcycle; mouse; oven; refrigerator; remote control; scissors; sofa; spoon; table; television; toilet; umbrella; watch (38) & Airplane; bed; chicken; goldfish; golf ball; hatchet; jeans; keyboard; knife; sink; toothbrush (11) \\
\bottomrule
\end{tabular}
\end{table}

\section{Additional visual-identity-disjoint split}
\label{app:identity_disjoint_split}

The original simple-split partitions separate individual stimuli but allow different variants of the same visual identity to appear in both training and validation. Specifically, 35 identities are shared between the original partitions, and 66 of the 80 validation stimuli have an identity also represented in training. We therefore provide an additional visual-identity-disjoint split of the same 400 stimuli, comprising 320 training stimuli covering 75 identities and 80 validation stimuli covering 20 identities, with zero identity overlap.

Table~\ref{tab:identity_disjoint_membership} lists the identities assigned to each partition. All color, position, and letter-case variants of a given identity belong to the same partition. The validation set contains letters C and S, digits 5 and 8, digit sequences 286 and 829, three geometric shapes, and eleven object categories. Color and position values remain represented in both partitions, so this protocol evaluates generalization to held-out visual identities while retaining shared attribute values.

The additional split can be reproduced by running the following command from the root of the accompanying GitHub repository:

\begin{tcolorbox}[colback=gray!8,colframe=gray!35,boxrule=0.4pt,arc=2pt,left=10pt,right=10pt,top=6pt,bottom=6pt]
\small\ttfamily
python scripts/generate\_simple\_identity\_disjoint\_split.py
\end{tcolorbox}

All reconstruction results reported in this paper use the original stimulus-level splits. The additional identity-disjoint split is provided as an optional evaluation protocol; baseline performance under this protocol has not been evaluated in the present study.

\newpage
\section*{NeurIPS Paper Checklist}

\begin{enumerate}

\item {\bf Claims}
    \item[] Question: Do the main claims made in the abstract and introduction accurately reflect the paper's contributions and scope?
    \item[] Answer: \answerYes{}
    \item[] Justification: The abstract and introduction state the paper’s contributions as a visual-free auditory-induced VMI paradigm, the AVID 5T fMRI dataset, a benchmark defined based on AVID, and baseline evaluations under the proposed protocol. The claims are limited to dataset construction, evaluation, and baseline characterization rather than model novelty.

\item {\bf Limitations}
    \item[] Question: Does the paper discuss the limitations of the work performed by the authors?
    \item[] Answer: \answerYes{}
    \item[] Justification: The paper discusses limitations in the Conclusion and Limitations section and Appendix A, including the Mandarin-speaking participant population, split imbalance, and the sensitivity of human neuroimaging data.

\item {\bf Theory assumptions and proofs}
    \item[] Question: For each theoretical result, does the paper provide the full set of assumptions and a complete (and correct) proof?
    \item[] Answer: \answerNA{}
    \item[] Justification: The paper does not present theoretical results, theorems, or formal proofs.

    \item {\bf Experimental result reproducibility}
    \item[] Question: Does the paper fully disclose all the information needed to reproduce the main experimental results of the paper to the extent that it affects the main claims and/or conclusions of the paper (regardless of whether the code and data are provided or not)?
    \item[] Answer: \answerYes{}
    \item[] Justification: The paper specifies the acquisition paradigm, stimulus splits, preprocessing, beta estimation, neural input definitions, metric suite, and baseline evaluation protocol. The dataset link provides detailed documentation on data organization and usage, and the accompanying code link provides scripts for data loading, metric computation, and baseline evaluation to support reasonable reproduction of the reported results.

\item {\bf Open access to data and code}
    \item[] Question: Does the paper provide open access to the data and code, with sufficient instructions to faithfully reproduce the main experimental results, as described in supplemental material?
    \item[] Answer: \answerYes{}
    \item[] Justification: The paper provides access to the de-identified AVID release, Croissant metadata, and code for data loading, metric computation, and baseline evaluation. The dataset and code links provide reviewer-accessible documentation and scripts for using the released data and reasonably reproducing the reported baseline evaluations.

\item {\bf Experimental setting/details}
    \item[] Question: Does the paper specify all the training and test details (e.g., data splits, hyperparameters, how they were chosen, type of optimizer) necessary to understand the results?
    \item[] Answer: \answerYes{}
    \item[] Justification: Sections 3–5 describe the dataset construction, participant task, stimulus splits, neural input settings, evaluation metrics, and baseline methods needed to understand the results. Appendix C provides additional baseline implementation details.

\item {\bf Experiment statistical significance}
    \item[] Question: Does the paper report error bars suitably and correctly defined or other appropriate information about the statistical significance of the experiments?
    \item[] Answer: \answerYes{}
    \item[] Justification: The paper reports participant-level means and sample standard deviations for reconstruction and attribute-level evaluation results (Tables \ref{tab:simple_attribute_results}, \ref{tab:variability_image} and \ref{tab:variability_text}). We also report paired Wilcoxon signed-rank tests for vividness-related differences in neural reliability (Table \ref{tab:vividness_neural_reliability}), statistical comparisons of fMRI-to-text decoding against shuffled-fMRI and matched-Gaussian controls (Table \ref{tab:caption_neural_validation}), and one-sided empirical permutation tests for MindEye1 reconstruction against shuffled-fMRI controls (Table \ref{tab:image_shuffle_validation}). The reported variability measures and statistical conclusions are specific to the corresponding experimental settings.

\item {\bf Experiments compute resources}
    \item[] Question: For each experiment, does the paper provide sufficient information on the computer resources (type of compute workers, memory, time of execution) needed to reproduce the experiments?
    \item[] Answer: \answerYes{}
    \item[] Justification: The paper reports the hardware used for the experiments and provides additional compute details in Appendix C, including GPU type, memory, and resources needed for the baseline evaluations.
    
\item {\bf Code of ethics}
    \item[] Question: Does the research conducted in the paper conform, in every respect, with the NeurIPS Code of Ethics \url{https://neurips.cc/public/EthicsGuidelines}?
    \item[] Answer: \answerYes{}
    \item[] Justification: The study was conducted in accordance with institutional ethical guidelines, with written informed consent from all participants. The released data are de-identified, and the paper documents intended uses and disallowed uses for AVID.

\item {\bf Broader impacts}
    \item[] Question: Does the paper discuss both potential positive societal impacts and negative societal impacts of the work performed?
    \item[] Answer: \answerYes{}
    \item[] Justification: The paper discusses potential positive impacts for VMI decoding and neuroprosthetic communication, as well as privacy and misuse risks associated with human neuroimaging data. Responsible-use constraints and disallowed uses are described in Appendix A.
    
\item {\bf Safeguards}
    \item[] Question: Does the paper describe safeguards that have been put in place for responsible release of data or models that have a high risk for misuse (e.g., pre-trained language models, image generators, or scraped datasets)?
    \item[] Answer: \answerYes{}
    \item[] Justification: The paper describes safeguards for releasing sensitive human neuroimaging data, including de-identification, random participant codes, removal of direct identifiers, defacing of anatomical images, and documentation of intended and disallowed uses.

\item {\bf Licenses for existing assets}
    \item[] Question: Are the creators or original owners of assets (e.g., code, data, models), used in the paper, properly credited and are the license and terms of use explicitly mentioned and properly respected?
    \item[] Answer: \answerYes{}
    \item[] Justification: The paper cites the existing datasets, models, software tools, and evaluation components used in the study. Appendix A.5 summarizes their roles and licensing or usage terms, and clarifies that the AVID license applies only to newly collected fMRI data, dataset metadata, and author-created materials.

\item {\bf New assets}
    \item[] Question: Are new assets introduced in the paper well documented and is the documentation provided alongside the assets?
    \item[] Answer: \answerYes{}
    \item[] Justification: The paper introduces AVID as a new dataset and provides documentation on dataset contents, structure, access, license, de-identification, limitations, and responsible use. The release also includes Croissant metadata and accompanying code for evaluation.

\item {\bf Crowdsourcing and research with human subjects}
    \item[] Question: For crowdsourcing experiments and research with human subjects, does the paper include the full text of instructions given to participants and screenshots, if applicable, as well as details about compensation (if any)? 
    \item[] Answer: \answerYes{}
    \item[] Justification: Sections 3.2 and 3.3 describe the participant task procedure, consent, compensation, withdrawal rights, and public data sharing after anonymization. Appendix A.6 provides the full participant-facing instructions.

\item {\bf Institutional review board (IRB) approvals or equivalent for research with human subjects}
    \item[] Question: Does the paper describe potential risks incurred by study participants, whether such risks were disclosed to the subjects, and whether Institutional Review Board (IRB) approvals (or an equivalent approval/review based on the requirements of your country or institution) were obtained?
    \item[] Answer: \answerYes{}
    \item[] Justification: The paper states that the study received institutional ethics approval and that all participants provided written informed consent in accordance with institutional guidelines. The participant procedure also describes withdrawal rights, compensation, anonymized data sharing, and MRI safety procedures.

\item {\bf Declaration of LLM usage}
    \item[] Question: Does the paper describe the usage of LLMs if it is an important, original, or non-standard component of the core methods in this research? Note that if the LLM is used only for writing, editing, or formatting purposes and does \emph{not} impact the core methodology, scientific rigor, or originality of the research, declaration is not required.
    \item[] Answer: \answerYes{}
    \item[] Justification: We used GPT-5.4 to translate Mandarin stimulus descriptions into English for documentation, model input, and text-based evaluation. Two authors manually reviewed and cross-checked the translations against the original Mandarin descriptions and revised them where necessary to preserve semantic accuracy. We also used Qwen3-VL-8B for attribute-level evaluation of reconstructed images.

\end{enumerate}

\end{document}